\documentclass[sn-mathphys,Numbered]{sn-jnl}

\usepackage{microtype}
\usepackage{graphicx}%
\usepackage{multirow}%
\usepackage{amsmath,amssymb,amsfonts}%
\usepackage{amsthm}%
\usepackage{mathrsfs}%
\usepackage[title]{appendix}%
\usepackage{xcolor}%
\usepackage{textcomp}%
\usepackage{manyfoot}%
\usepackage{booktabs}%
\usepackage{algorithm}%
\usepackage{algorithmicx}%
\usepackage{algpseudocode}%
\usepackage{listings}%
\usepackage{multicol}
\usepackage[russian,french,english]{babel}
\usepackage[utf8]{inputenc}
\usepackage[T2A,OT1]{fontenc}
\usepackage{tempora} 
\usepackage{epsfig}
\usepackage{calc}
\usepackage{amstext}
\usepackage{multicol}
\usepackage{pslatex}
\usepackage{csquotes}
\usepackage[textsize=tiny]{todonotes}
\usepackage{url}
\usepackage{tabularx}
\usepackage{standalone}
\usepackage{blindtext}
\usepackage{hyperref}
\usepackage{longtable}
\usepackage{nicefrac} 
\usepackage{subcaption}
\theoremstyle{thmstyleone}%
\theoremstyle{thmstyletwo}%

\theoremstyle{thmstylethree}%

\newcommand{\nocomment}[1]{#1}
\newcommand{\COMMENT}[1]{}

\newcommand{\textcite}[1]{\citep{#1}}

\newenvironment{itemizerCompact}{\vspace{-1mm}
  \begin{itemize}
    \setlength{\itemsep}{2pt}
    \setlength{\parskip}{0pt}
    \setlength{\parsep}{0pt}
  }
{ \end{itemize}
  \vspace{-1mm}  }

\newenvironment{enumeratorCompact}{\vspace{-1mm}
  \begin{enumerate}
    \setlength{\itemsep}{2pt}
    \setlength{\parskip}{0pt}
    \setlength{\parsep}{0pt}
  }
{ \end{enumerate}
  \vspace{-1mm}  }

\begin{document}

\COMMENT{
\title[Is implicit assessment of language learning during practice as accurate as assessment through testing?]{Is implicit assessment of language learning during practice as accurate as assessment through testing?}
}

\title[Implicit assessment of language learning during practice as accurate as explicit testing]{Implicit assessment of language learning during practice as accurate as explicit testing}


\author*[1,2]{\fnm{Jue} \sur{Hou}}\email{firstname.lastname@helsinki.fi}

\author[1,2]{\fnm{Anisia} \sur{Katinskaia}}

\author[1,2]{\fnm{Anh-Duc} \sur{Vu}}

\author*[1]{\fnm{Roman} \sur{Yangarber}}

\affil[1]{\orgdiv{Department of Digital Humanities}}

\affil[2]{\orgdiv{Department of Computer Science}}

\affil{\orgname{University of Helsinki}, \orgaddress{\country{Finland}}}

\abstract{
  
Assessment of proficiency of the learner is an essential part of Intelligent Tutoring Systems (ITS).
We use Item Response Theory (IRT) in computer-aided language learning for assessment of student ability in two contexts: in test sessions, and in exercises during practice sessions.

Exhaustive testing across a wide range of skills can provide a detailed picture of proficiency, but may be undesirable for a number of reasons.  Therefore, we first aim to replace exhaustive tests with efficient but accurate adaptive tests.
We use learner data collected from exhaustive tests under imperfect conditions, to train an IRT model to guide adaptive tests.
Simulations and experiments with real learner data confirm that this approach is efficient and accurate.

Second, we explore whether we can accurately estimate learner ability directly from the context of practice with exercises, without testing.  We transform learner data collected from exercise sessions into a form that can be used for IRT modeling.  This is done by linking the exercises to {\em linguistic constructs}; the constructs are then treated as ``items'' within IRT.

We present results from large-scale studies with thousands of learners.
Using teacher assessments of student ability as ``ground truth,'' we compare the estimates obtained from tests vs. those from exercises.  The experiments confirm that the IRT models can produce accurate ability estimation based on exercises.

}




\maketitle

\section{Introduction}
\label{section:introduction}

An important goal in intelligent tutoring systems (ITS) is to support the learner's progress through {\em personalized} tutoring.  ITSs have been shown to be effective in various subject domains,~\cite{Ritter2007,Arroyo2014,klinkenberg2011computer}.
We present work in a sub-area of ITS, namely on computer-aided language learning (CALL), used in a real-world learning environment.  We present experiments with the CALL system 
Revita, \cite{2023:eacl:constructs,katinskaia:2018-lrec:revita}.
This work is the result of a collaborative international effort, involving teachers who use the system in their curricula at several universities; the approach is targeting students who have passed beyond the beginner level.

The main aspect of Revita's 
approach to CALL is that the intelligent tutor supports the student during {\em out-of-class practice}.  The student may attend lectures, receive various learning content from the teacher, etc.; the tutor engages the student in practice, to leverage the student's time outside the classroom toward improving mastery.

A central requirement for personalized tutoring in ITS and CALL is {\em accurate assessment} of the current proficiency of the learner.  Assessment is needed for two main reasons: the ``external'' and the ``internal.''  Externally, assessment is needed to communicate to the learner---and to the teacher---what the learner has/has not mastered.  This supports planning future lectures, individual attention, etc.
More crucially, internally, the assessment drives the choice of exercises presented to the learner during practice.  The ITS contains an Instruction Model, which is the component responsible for selecting the most appropriate exercises for the learner, given the learner's current state of proficiency.

\begin{figure}[t]
  \centering
  \center
  \includegraphics[trim=10mm 49mm 40mm 39mm, clip, width=.75\columnwidth]{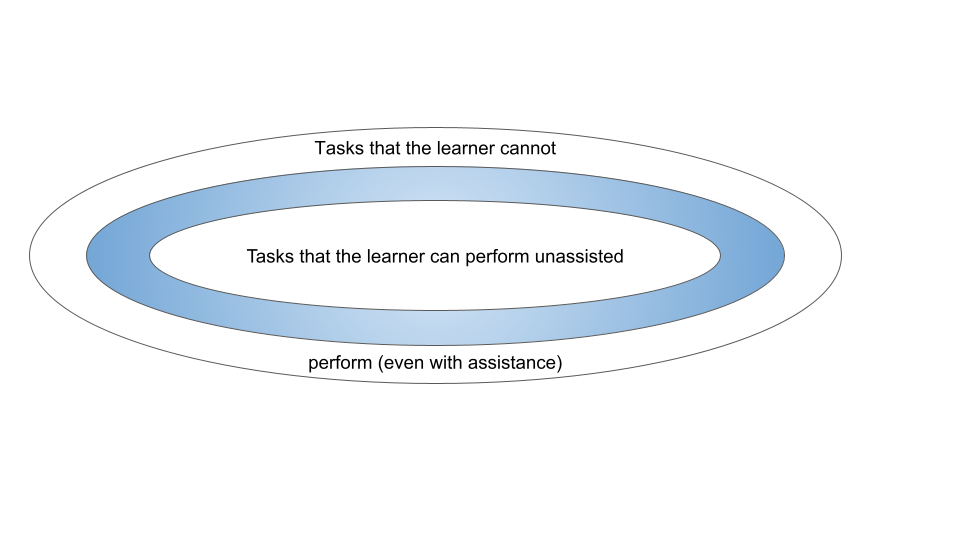}
  \caption{{\em Zone of proximal development}: the blue area---tasks that the learner can perform with some assistance are those that the learner is most prepared to learn next.}
  \label{fig:zpd}
\end{figure}

In the widely-accepted Vygotskian theory of educational psychology, the {\em Zone of Proximal Development} (ZPD) is defined as those  skills in the learning domain, which the learner is best prepared to learn next, see Figure~\ref{fig:zpd}~\cite{vygotsky1978mind,vygotsky2012thought}.  Skills falling outside the ZPD are either already mastered, or they are too far beyond the learner's reach at present.  If the exercises offered by the tutor to the learner focus excessively on skills already mastered---it will bore the learner.  If we focus excessively on skills far out of reach---it will frustrate the learner.  In either case, the learner will be demotivated and is likely to quit learning.
Therefore the ITS should concentrate its efforts on finding the ZPD, to make sure that most of the time the exercises are exactly at the right level.
Identifying the ZPD requires accurate assessment of the learner's ability.

We must distinguish from the outset two different contexts (environments) in which the learner can work with the tutor: {\em test} sessions vs.~practice sessions with {\em exercises}.
In this learning setting, we will explore three types of assessment: (A) assessment via exhaustive test, (B) assessment via {\em adaptive} test, and (C) assessment via {\em exercises} done by the learner during practice sessions.

One ``traditional'' approach to assessment involves {\em exhaustive} testing,
%
where the student answers a long list of test questions (items).
The test typically follows a template identical for all learners.
The advantage of an exhaustive test is that it can give the teacher 
a detailed picture of the student's ability.
However, it may have several weaknesses: 

\begin{enumeratorCompact}
  
  \item[(i)] Exhaustive tests can take a long time, which may be stressful for students.
Testing may give a skewed assessment, since the testing context itself impacts the outcome.  Students may perform better or worse in a testing environment.  Strictly, a test may measure the student's ability to {\em perform on a test}---rather than the target ability.
In light of this, consciously or not, students may train for the test, rather than for the ultimate skills.

\item[(ii)] The ``brute-force'' exhaustive approach---testing each skill separately and repeatedly---should not be necessary, since the skills are not independent: the skill space has an internal structure, for example, in terms of {\em prerequisite} relations.
This means that many questions in an exhaustive test may be redundant.  However, these relations between the skills may be latent and not explicitly known.

\item[(iii)] Most importantly: from the learning perspective, testing is not an activity that promotes learning {\em per se}, as does reading, practicing with exercises, etc.
\nocomment{Strictly, testing is not a learning activity.  For example, during a test, the learner is not informed whether her answer to any given question is correct.  Exercises are designed as a learning activity, with immediate feedback for the student's answers, and elaborate hints that gradually lead the learner toward finding the correct answer on her own.}
\end{enumeratorCompact}

Concerning redundancy, if the examinee demonstrates very strong overall proficiency, then most ``easier'' questions are redundant---they will yield no useful information.  A more efficient solution is to use the structure of the learning domain and the inter-dependencies among the skills to adjust the sequence of questions adaptively.

We have conducted large-scale studies with several thousand students, and collected large amounts of learner data from tests and exercises.\footnote{All data (anonymized) is released to the community through Github for verifiability and further research.}
For test assessment (A) and (B), we work with a large bank of question items designed by experts in language pedagogy.
%
For exercise assessment (C), we work with arbitrary texts that the users (teachers and learners) upload to the system.  Revita 
generates exercises based on the chosen text.  Each exercise is linked to a set of linguistic constructs.  The taxonomy of linguistic constructs is created and validated by experts in language pedagogy, and is continually evolving over time.  The eventual goal is to capture many (hundreds) of constructs that are essential to the mastery of the grammar of the language.\footnote{At present, Revita 
is used by university learners of Finnish and Russian; since more data has been collected from a larger number of sites for Russian, the experiments presented in the paper will focus on Russian.} 

We aim for accurate monitoring of the learner's proficiency and progress during exercise sessions over time, to offer better personalized exercises and feedback to the learner.
The collected learner data is imperfect in several respects (Section~\ref{section:data-challenges}).

We present work on assessment models trained on test data as well as models trained on exercise data using Item Response Theory (IRT).
The research questions we aim to answer are:
\begin{itemizerCompact}

\item {\bf RQ1}: Can we build a reliable {\em adaptive} model of learner ability---from the learner data? That is, a model that is much quicker than exhaustive testing, but equally accurate.

\item {\bf RQ2}: Does imperfect learner data still provide robust assessment of learner ability?

\item {\bf RQ3}: How do estimates of ability from IRT trained on learner data compare with estimates of ability based on question difficulty assessed {\em manually} by teachers?

\item {\bf RQ4}: Can we reliably model learner ability based on the learner responses to exercises?

\end{itemizerCompact}

The paper is organized as follows.
Section~\ref{section:prior} introduces relevant prior work.
Section~\ref{section:data} describes the test item banks and the data collected from the learning platform.
Section~\ref{section:method} describes in detail the methodology training and tuning of the IRT models. 
Section~\ref{section:experiments} presents the experiments and results on assessment of learner ability. 
Section \ref{section:conclusion} concludes with the current directions of research.

\section{Prior work}
\label{section:prior}

\paragraph{Learning analytics:}
Several approaches for modeling learning are currently in use.
Bayesian Knowledge Tracing (BKT) attempts to model student proficiency as a latent variable in a Hidden Markov Model, which is updated by observing learner responses to test items~\cite{corbett1994knowledge}.
Surveys of BKT research may be found, e.g., in \cite{2021:liu:survey-BKT,2017:pelanek:bayesian-knowledge-tracing}.
BKT variants train separate parameters for initial knowledge and learning; they also model ``guess'' and ``slip'': {\em guess} refers to when the student answers a question correctly although she has not mastered the skill; conversely, a {\em slip} happens when the learner has mastered a skill, but gives a wrong answer accidentally.
BKT models may estimate guess and slip for each skill; or may compute them each time a student attempts to answer a new question, based on learned models of guess and slip, e.g.,~\cite{baker2008improving}. We turn to these topics in Section~\ref{section:slip_exp-introduction}.

\nocomment{
  Knowledge Space Theory (KST), models the domain as a ``knowledge space''---a graph of knowledge {\em states}, in which a student may be at the present time~\cite{doignon2012knowledge}.  Each state contains a subset of the skills in the domain, mastered by the student so far.
  The student has mastered the domain once she reaches the state containing all skills.  KST models not only the student knowledge, but also learning paths, starting from the empty set toward the full set of concepts.  One approach to building a knowledge space is Formal Concept Analysis (FCA)~\cite{ganter2012formal}.
}

Item Response Theory (IRT) is a popular approach to modeling the student's state of ability,
\cite{van2013handbook}.
IRT is applied in many settings (besides ITS), including stress testing, psychological and medical testing, etc.
Depending on the specific application domain, the latent trait under consideration can encompass various factors such as anxiety, neurosis, authoritarian personality, and so on. Several studies have focused on the application of Item Response Theory (IRT) in the field of language learning. For instance, these studies have involved modeling language proficiency using test data~\cite{alavi2021examination}, validating the reliability and difficulty of language tests using IRT~\cite{sudaryanto2019applying}, and measuring the difficulty of Thai language topics for learners~\cite{runzrat2019applying}. Furthermore, Polat~\cite{polat2022comparison} conducted a comparison of language proficiency performance measures based on test data using both IRT and Classical Test Theory (CTT), with IRT demonstrating superior capabilities in modeling language skills.
We apply IRT to language learning, where the latent trait is the learner's ability, or language proficiency.
IRT has an information-theoretic basis; in that it is related to Elo rating systems of ability~\cite{elo1978rating}---the Elo formulas have been adapted to the context of ITS,~\cite{pelanek2016applications}.
%
The language-learning domain is much more complex than many domains where IRT is used, because the learning {\em constructs} to be mastered by the learner are relatively numerous,
and have complex relationships among them.  A linguistic construct corresponds to a particular narrowly defined skill or topic that is to be mastered by the learner, e.g., a specific
grammatical concept, such as a particular type of verbal inflection.

{\bf Adaptive test:} 
The goal of computerized adaptive testing (CAT) is to achieve an accurate score with a {\em reasonably small number} of questions~\cite{2010:linden:elements-adaptive}.  This makes it critically important to find good termination criteria~\cite{linden2000computerized}.



Most approaches proposed for choosing a termination criterion focus on finding a criterion for a {\em classification} problem, \cite{thompson2011termination}.
This may make sense when there are well-defined boundaries between proficiency levels.  In language learning, we use the CEFR scoring system,\footnote{https://www.coe.int/en/web/common-european-framework-reference-languages/} where no clear boundaries between ratings are defined, therefore we view the problem in terms of {\em regression}.

Further prior work is discussed in the following sections {\em in situ}, such as work on termination criteria in Section~\ref{section:termination-criteria}, etc.

\section{Data}
\label{section:data}

This work builds on learner data collected through a collaborative effort with language teachers at several universities, with learners at various levels on the CEFR scale, ranging from A1 to C2,~\cite{little-2007-CEFR}.
The platform collects learner data from several languages and multiple contexts: grammar exercises, vocabulary exercises, tests, etc.  Here we focus on data collected from tests and grammar exercises from learners of Russian as a second language (L2).

\subsection{Learner data from tests}

Several thousand students took many exhaustive tests.
Each test session consists of items sampled from a bank of 3390 multiple-choice questions.  The question bank is compiled by experts in language pedagogy. 
Each question is linked to one of a set of linguistic constructs (140 at present); which constructs are involved in each question is also determined by the experts.  Each teacher can define a template, selecting which constructs to test (or use a ``default'' template with 300 questions).
All of these items are dichotomous: a learner's response to an item is either correct or wrong.\COMMENT{This dataset, fully anonymized, is released for use by the research community, through Github.}

The experiments we describe use a data set of over 750 000 responses to test items from approximately 1800 learners.  For each question, we record to which construct the question relates, whether the answer was correct, and the timestamp.

\paragraph{Manual assignment of difficulty and proficiency:}
For each item (test question)
the pedagogy experts assigned a {\em difficulty} score on the CEFR scale for the linguistic constructs linked to the item.  Also, the teachers assigned a CEFR level of {\em proficiency} to over 200 students, based on overall performance in their language courses.
The estimates of difficulty made by the experts for many of the questions in the item bank were considered questionable by other experts.
In our experiments, we aim to evaluate the quality of the judgements of difficulty and proficiency by comparing them with the estimates from models trained on learner data ({\bf RQ3}).

\subsection{Learner data from exercises}


All exercises are automatically generated by the CALL system, Revita, 
based on authentic texts chosen by the learners from arbitrary sources on the web. The system creates a number of exercise types; here we focus on fill-in-the-blank (``cloze'') and multiple-choice exercises.  In a cloze exercise, the system hides certain words or phrases, and the learner receives the lemma (dictionary form) of the hidden word or phrase as a hint.  The learner's task is to insert the correct surface forms, based on the context of the cloze.
In a multiple-choice exercise, the learner is given several options to choose from, the options are generated automatically.

The learner can make multiple attempts to answer each exercise.  In case of an incorrect answer, on subsequent attempts the system gives hints, which gradually guide the learner toward the correct answer. The hints start out as more general and become progressively more specific on subsequent attempts.
Each exercise is attached to one or more linguistic constructs.
The system analyzes the student's response, the requested hints and the number of attempts to answer to compute the performance of each linguistic construct.  In general, a student's answer may be correct with respect to some constructs, but wrong with respect to others; for example, the student may have entered a correct tense for a verb but an incorrect person.

We have collected 214K exercises, done by approximately 1500 students.
The exercises are linked to over 200 linguistic constructs, \cite{2023:eacl:constructs}. 
We have over 50 students, who have done over 100 exercises and who have their CEFR levels assessed by their teachers.
We evaluate the quality of the model trained on all students, by comparing the estimated proficiency of these students with their teacher-assessed CEFR levels.

\COMMENT{??? ??? experiments with manually assigned CEFR difficulties ?}

\subsection{Challenges relating to the data}
\label{section:data-challenges}

\paragraph{Tests:} Although the exhaustive test is developed by experts in language pedagogy, it is problematic in several aspects.  First, the test is too long.  Students reported being stressed and frustrated by the end of the test.  Second, the test originally allowed only 15 seconds per question, which is too short, especially when the examinee is at the lower CEFR levels.  Because of these problems, the data which we collected from the exhaustive test may give a skewed reflection of students' true proficiency.
This poses a challenge---can robust models for testing be trained on this imperfect learner data ({\bf RQ2}) ?

\paragraph{Exercises:} Evaluating the exercises is also very challenging. The challenge comes with 
%
assignment of credit and penalty to the student answers. 
%
%
In terms of assigning credit and penalty, a major challenge is that language constructs are not directly judged, as test questions are. Each test question corresponds in a clear way to one {\em item}, and assigning credit/penalty is straightforward for test questions: it is unambiguous and there is a clear judgment of the answer---correct or wrong.
In the case of exercises, the credit standard is not clear because the link from exercise to linguistic constructs is one-to-many. This requires a more sophisticated approach to credit and penalty assignment. We will discuss our approach in more detail in Section~\ref{section:method}. Overall, these challenges pose a similar question: can we use the data (possibly noisy) from exercises to construct robust models ({\bf RQ4})?

\section{Method}
\label{section:method}

\subsection{Item Response Theory}
\label{section:IRT}

Item Response Theory (IRT) provides ways to evaluate and compare the difficulty of
question items and the proficiency of a
learner---which in our setting should be mapped to the CEFR levels.

{\bf 3PL model:} Since our test items are multiple-choice questions with one correct answer, we focus on the 3PL variant (three-parameter logistic model) of IRT~\cite{van2013handbook}.
In 3PL, we model the probability that a student $s$ whose current ability estimate is $\theta_{s}$ will give a correct answer to $Q_{i}$--- Question item $i$.  The probability function is expressed as:
%
\begin{align} 
  & P( \theta_s, Q_i )
  = {c_i} + (1 - {c_i}) \cdot \frac{1}{1 + exp( -a_i ( \theta_s - b_i ))}
    \label{eq:3pl}
\end{align}
%
where the parameters---the properties of $Q_i$---are:
\begin{itemizerCompact}
\item $a_i$: discrimination factor,
\item $b_i$: estimate of difficulty,
\item $c_i$: probability that a random guess is correct.
\end{itemizerCompact}

{\bf Information Function:} 
IRT defines two types of information functions: {\em item} information and {\em test} information.
Item information measures the amount of information a question $Q_i$ yields, based on the
learner's current ability estimate $\theta_{s}$.
The Information function is used during the adaptive test to select the most informative item, for the given value of ability  $\theta_{s}$.
It is computed as:
%
  \begin{equation}
    I(\theta_s, Q_i) = 
    a_i^2\frac{1 - P(\theta_s, Q_i)}{P(\theta_s, Q_i)}\left [ \frac{P(\theta_s, Q_i) - c_i}{1 - c_i}\right ]^2
  \label{eq:info}
\end{equation}
%
{\em Test information} is the sum of information over all items.  It provides a measure of the informativeness of the test as a whole; i.e., it measures how accurate the test result will be.

\subsection{Test Analytics}

We next describe the experimental setup: evaluating the effectiveness of IRT under various conditions.

\subsubsection{Test simulations}

We simulate the adaptive test, to check the effectiveness of IRT trained from learner data---for testing future unseen learners.

{\bf The Simulation procedure:} 
iteratively selects successive question items as follows:
\begin{enumeratorCompact}
\item Choose $\theta_{true}$---this is the {\em true} ability of the learner (hidden from the model).

\item Initialize $\theta_{n}$, an estimate of ability $\theta_{true}$.  For time step $n=0$, the
  initial $\theta_{0}$ is drawn from the normal distribution with $\mu = 0$,
  $\sigma = 0.5$; i.e., initially the student is ``about average.''
  
\item Item selection: given the current value of ability $\theta_{n}$, find the next
  ``best'' question---the most informative item $Q_{n}$ according to
  Equation~\ref{eq:info}.
  \label{step:item-select}

\item ``The learner answers'' the selected question at time $n$:
  in the simulation, the probability of a correct answer is sampled using the {\em true} ability $\theta_{true}$ and
  Equation~\ref{eq:3pl}.
  \label{step:sample-answer}

\item Update the estimate of ability $\theta_{n+1}$, based on the answers observed and the
  parameters of all items $Q_i$ answered so far, for $i=0,1,...n$.
\item Repeat steps 3--5, until $\theta_{n}$ converges, according to the termination
  criterion (see below).
\end{enumeratorCompact}

It is important to note that although $\theta_{0}$ is initialized randomly,
in practice, for learners with {\em similar true} proficiency,
this adaptive test procedure will quickly converge to the {\em same sequence} of questions: after a small number of random questions at the beginning, the model will offer exactly the same questions.
We want to avoid this kind of repetitiveness in testing.  The goal for the test is to be not only accurate, but also {\em varied}.  If we offer almost the same set of questions in different test sessions, then, intentionally or not, the examinees will come to memorize those ``fixed'' questions, which will eventually skew the outcome estimates.

Further, over time we want to collect learner responses for {\em many different} items in the item bank, to be able to train better IRT models in the future.\footnote{In real life during the adaptive test (not in simulations), for 10\% of items, we select randomly from the items for which we have the {\em fewest} learner responses.  Even if these are considered low-information items, we want to obtain richer data in our item bank, with more data collected for all items, rather than having a skewed bank, where some items have many responses, and others nearly none.}
For these reasons, in practice, rather than selecting the single {\em most} informative question at time step $n$, we sample a question randomly from the {\em top 5} most informative questions.
This is done during adaptive tests with real-world students; as a faithful reflection of the real-world behavior, we use the same policy in simulations.

\subsubsection{Slip and Compensation}
\label{section:slip_exp-introduction}

{\bf Slip:} Although Equation~\ref{eq:3pl} includes the notion of a lower bound for random
guess,
the simulation procedure does not include the possibility of a slip, which may happen {\em randomly}, even if the learner knows the correct answer.  Therefore, we introduce an extra step for slip when a {\em correct} learner answer has been sampled from Equation~\ref{eq:3pl}, in step \ref{step:sample-answer}.

Research has shown that a participant's willingness or capability to attend to the sequence of test questions is an important contextual factor: motivated participants may process information more deeply~\cite{petty1986elaboration,haugtvedt1989need,petty1995elaboration}.  This suggests that a slip is related to the examinee's attitude toward the test, and it may be unrealistic to estimate the probability of a slip $\epsilon_{slip}$.
In our experiments, we expect $\epsilon_{slip}$ to be fairly low---we assume that the student is trying to concentrate on the test to the best of her ability.  In our simulations, we use $\epsilon_{slip} = 0.05$.

{\bf Exploration:} We expect that the presence of slips will {\em slow down} the adaptive test process.  The mechanism of IRT will try to compensate for the ``drop'' in ability estimate $\theta$ and will select questions at a lower difficulty, which will lead to a longer test.  To speed up the process, we introduce a capacity for {\em exploration} before step~\ref{step:item-select} (item selection).  With probability $\epsilon_{expl}$, we add a factor $\alpha$ to $\theta_{n}$, according to its trend over the previous several time steps: if the trend is downward, $\alpha$ is negative, and if it is upward, $\alpha$ is positive.
In our simulations, the probability of exploration $\epsilon_{expl} = 0.2$.

Implementing exploration requires a reliable estimation of the {\em trend} in $theta$ over the recent $N$ iterations. Therefore, we do not apply exploration at the beginning of the test (since it takes some time to establish a trend).

On the other hand, we see, for example, in Figure~\ref{fig:simu_artificial} that after 25 (or more) iterations the adaptive test may already be quite close to the true target estimate of ability.

Therefore, as the trend becomes flat, we believe we should stop exploration.
The goal of exploration is to speed up the convergence, not to delay the convergence.
But we expect that conducting exploration throughout the whole test may ``confuse'' the IRT, as exploration will select questions that are harder than the true level, if the overall trend is positive, or easier than the true level, if the overall trend is negative.
This may lead to extra loss or gain in ability estimation and, therefore, prolong the test. This is not desirable.

We explore various values of parameters $\alpha$ and the range of exploration $N_{exp}$.
\COMMENT{
Due to the exploration factor $\alpha$, questions that are higher or lower than the
learner's true proficiency will be selected later during the test.  This will confuse the
IRT model since it will mostly cause extra loss or gain in estimation and, therefore,
extend the test.  This is not desirable either.

We currently apply exploration between 10th to 30th iteration.  These limits are based on
observations from our experiments in from Section~\ref{section:experiments}.  We wait
until after the first 10 iterations, because we want to observe some trend in ability
estimates; we stop exploring after 30 iterations, because, we see, for example, in
Figure~\ref{fig:simu_artificial} that after 25 iterations the adaptive test may already be quite close to the true target estimate of ability.}
In the experiments, the sources of randomness are: random initialization, random item selection, slip, and exploration.

{\bf Warm-up:} In addition to normal slip, which may happen randomly throughout the test, we observe that in practice\COMMENT{??? how often :/  should we have statistics?} more slips occur at {\em the beginning} of a test.
Such aberrant responses at the early stage have a strong negative impact on the final ability estimate, which is then significantly underestimated as a result.
This problem has been acknowledged and studied in prior work, e.g.,~\cite{rulison2009ve}. In practice, an early aberrant response may be frequent due to a variety of external factors. For example, a student may be unfamiliar with the (online) test environment, may be stressed by taking the test, etc.

In avoid this problem, we introduce a {\em warm-up} phase. 
During warm-up, we follow the same test procedure as usual, except that wrong responses are ignored, and not fed into the IRT model. The number of questions in the warm-up phase is set to 10.  We choose this number because we observed most of our users tend to have a better correct rate after 10 questions.  Therefore, we suppose that most users are settled in after 10 questions.  The warm-up phase is designed to help the learner get familiar with the test environment and to remove the effect of early aberrant responses.  After the warm-up, the regular adaptive procedure is applied, and all responses, correct or wrong, are used for ability estimation.  We show the results of the simulation of warm-up in Section~\ref{section:experiments}.

\subsubsection{Termination Criteria}
\label{section:termination-criteria}

We explore how termination criteria affect the estimation at the end of the simulation.
We mostly follow the termination criteria evaluated in \cite{babcock2009termination}, although our context and setup are somewhat different.
\textcite{babcock2009termination} performed a comprehensive evaluation over four criteria: fixed-length, Standard Error of Measurement (SEM), change in $\theta$, and remaining minimum information.  They also evaluate combinations of these criteria.
Inspired by this work, we apply the first three criteria, as the fourth one did not perform well in their evaluation.

\vspace{1mm}\noindent
{\bf Fixed-length:} The simplest approach is to terminate a simulation after a fixed number of iterations.
This criterion can give a general idea of the performance of the test, when the rest of the test does not offer more information and the predicted $\theta$ is reasonably close to the true level.  This also serves as a baseline for the other termination criteria.

\vspace{1mm}\noindent
{\bf SEM:} This variable-length termination criterion depends on the {\em confidence} of $\theta$.  The iterations will go on until the test reaches a level of confidence, which is measured by standard error---the {\em variance} of the ability estimate in the 3PL model.  It can be estimated as the reciprocal of the test information function at the current ability estimate $\theta$:

  \begin{equation}
    SEM(\theta) = \sqrt{1 / \sum_i{a^2\frac{1-P(\theta, Q_i)}{P(\theta, Q_i)}\left[\frac{P(\theta, Q_i) - c}{1-c}\right]^2}}
  \end{equation}

\vspace{1mm}\noindent
{\bf EarlyStop:} This variable-length termination criterion is based on the {\em change} in $\theta$.  We expect $\theta$ to change more at the beginning of the test, when each question offers much information about the examinee, and less at the end. 
Less change in $\theta$ suggests a better estimation of $\theta$.  In our simulations, we use the idea of EarlyStop~\cite{morgan1989generalization} from machine learning (ML) training procedure, i.e., we assume $\theta$ is converged when $|\theta_n - \theta_{n-1}| < \delta$ for $N$ time steps, for some $\delta$ and $N$.

\vspace{1mm}\noindent
{\bf Metrics:} We use four metrics as indicators of performance of different parameter setups:
\begin{itemizerCompact}

\item Mean number of iterations: the length of the test, measures the efficiency of the test.

\item Standard Deviation (SD) of the number of iterations: measures how stable a test setup is.

\item Mean Absolute Error (MAE): between the predicted ability $\theta$ and true ability
  $\theta_{true}$.  Low MAE indicates that the test is more accurate.

\item Standard Deviation (SD) of error: this indicates how the outcome is affected by
  different termination criteria.

\end{itemizerCompact}

Mean Absolute Error and SD of absolute error are affected by the termination criteria and
also by the randomness introduced into our simulations: by slip, exploration, and sampling
from the top-5 most informative questions.  Therefore, we also use these metrics when
evaluating the effect of slip and exploration, in Figure~\ref{fig:exp_slip}.

\subsection{Exercise Analytics}
\label{sec:exercise-analytics}


Alongside with experiments above on adaptive and exhaustive test data, as mentioned in the Introduction, we aim to explore how much information about the student's ability can be inferred from her performance of exercises---collected over practice sessions, without explicit testing.  A good teacher can form reliable estimates of the student's ability without explicit testing; the AI tutor aims to do the same.

The key question is: to apply IRT, we need a notion of an ``item'' in the exercise setting.  The item {\em cannot} be an exercise---since all students work with different texts, chosen by them individually, the exercises they receive are also different!

To model the exercise learner data, we propose to consider the {\em linguistic constructs} as the items for the IRT model.
To calculate the performance of students on the constructs, we need to analyze the learner's answers and compare them to the expected answers.
Each exercise involves more than one construct in general.
When a learner answers an exercise, she may get some of the constructs correct---e.g., the {\em tense} of the verb---and other incorrect---e.g., the {\em person} of the verb.
We assign credit for the constructs that the learner answered correctly, and penalty for constructs for which the expected answer and learner's answer differ.

Note: the hints and feedback that the learner requests are also linked to constructs:
``use perfect tense,'' ``use second person,'' etc.
We also penalize those linguistic constructs that are linked to hints that the learner requested:
since the learner has asked for a hint, we can assume that she needed help---has not yet mastered the corresponding construct.
Based on the counts of credits and penalties for each linguistic construct, we define the performance of every student $S$ on every construct $C$ as the rate of correctness of $S$'s answers on $C$.

We assign the guess factor as $0.0$ to ``cloze'' exercises, and $0.25$ to ``multiple-choice'' exercises.


\section{Experiments and results}
\label{section:experiments}

The main question for the simulations is whether the IRT models trained on our learner data work, and if they work, what do the simulations reveal?
We explore these questions next.

\subsection{Experiments with test data}

\subsubsection{Estimation of student ability}


\begin{figure*}
  \begin{minipage}{0.49\linewidth}
    \centering
    \center
    \includegraphics[trim=5mm 0mm 8mm 0mm, clip, width=\columnwidth]{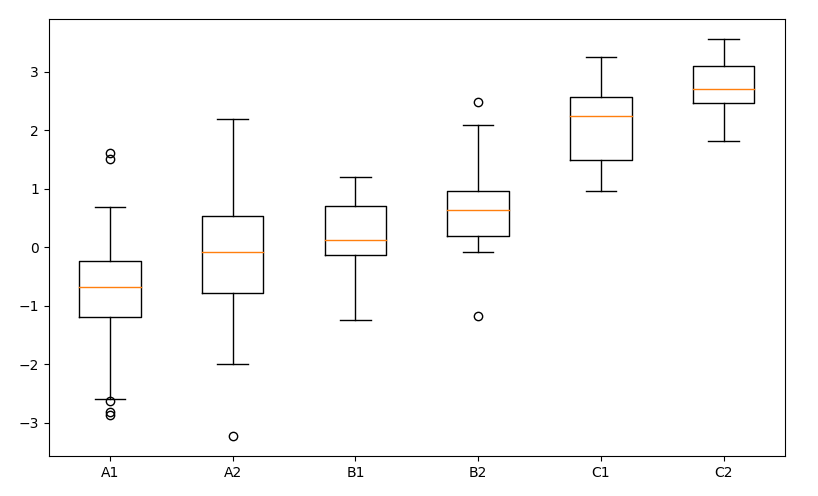}
    \caption{Simulation on real learner data with manually labelled CEFR levels of learners.
      X-axis---the 6 CEFR levels; Y-axis---ability estimate.}
    \label{fig:simu_real}
  \end{minipage}\hfill
  \begin{minipage}{0.49\linewidth}
    \center \includegraphics[trim=2mm 0mm 3mm 0mm, clip, width=\columnwidth]{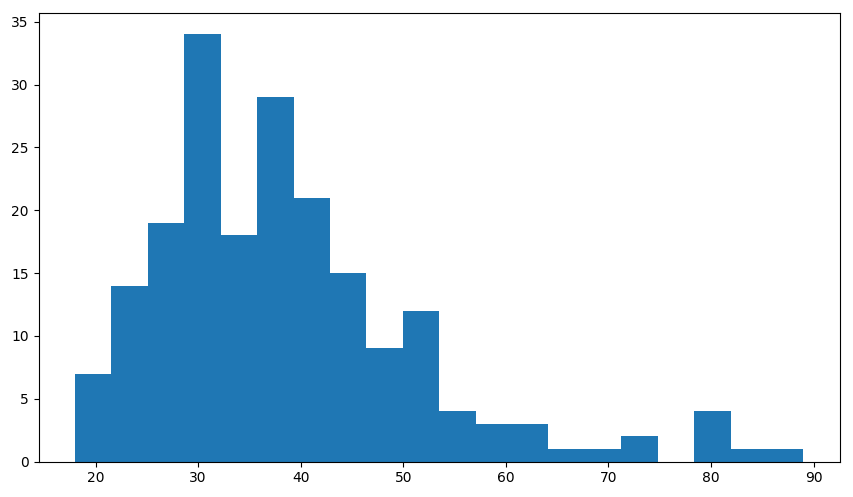}
    \caption{Distribution of length of adaptive test in real learner simulations in
      Figure~\ref{fig:simu_real}.  X-axis---length of test; Y-axis---number of tests in each bin.}
    \label{fig:simu_real_distrib}
  \end{minipage}
\end{figure*}

To evaluate the ability estimates that the IRT-based test provides in real life, we conduct a simulation with test data from 200+ \textbf{real} learners.  These learners were assigned by the teachers manually to a CEFR level (A1 to C2), and each learner took the exhaustive test at least once  (300+ questions).

The exhaustive test follows a {\em fixed} template: the questions in the test are not necessarily the most informative---as the IRT-based adaptive test would select.  Therefore, in the simulation, rather than selecting the most informative question {\em overall}, we select the most informative question only from among those for which we have an actual response from the learner.
Lastly, we limit the length of this simulation to a maximum of 100 steps.\footnote{A heuristic choice---maximum tolerated length of a ``short'' test, inspired in part by the artificial simulations in Fig.~\ref{fig:simu_artificial}.}

Figure~\ref{fig:simu_real} and Figure~\ref{fig:simu_real_distrib} show the
results with {\em real} learner data.
Figure~\ref{fig:simu_real_distrib} shows the distribution of the test lengths---the vast
majority of the tests converge in 60 questions or less.
Figure~\ref{fig:simu_real}, shows very strong correspondence
between the IRT model's prediction and the teachers' CEFR judgements.
%
This matches with the results from the artificial simulations, and confirms that the IRT-based adaptive test is effective---much shorter test length---and accurate---its predictions of ability correspond well with the teachers' judgements ({\bf RQ1}).

Note that the range of the {\em observed} ability scores over the students in this group
is $[-3, 3]$.
\COMMENT{Note that if we split the range of
  observed ability scores (between $-1$ and $3$) into six boxes of approximately equal
  size, that would allot about $.66$ 
  units of ability for each CEFR level.}


%

%
Although examinees with a higher CEFR score have a higher ability estimate, Figure~\ref{fig:simu_real} shows a less clear boundary between some levels, especially levels A2 and B1 show some overlap.  This may be because our examinees are not evenly distributed across the CEFR levels.  At levels B1 and B2 in our dataset has fewer learners---23 and 14 examinees, respectively;  larger numbers of examinees at A1 and A2, with 87 and 55 examinees, respectively.  We can expect clearer boundaries between levels when more data are collected and more examinees are labeled with CEFR levels.

\begin{figure*}
  \begin{minipage}{0.49\linewidth}
    \centering
    \center
    \includegraphics[trim=4mm 0mm 4mm 0mm, clip, width=\columnwidth]{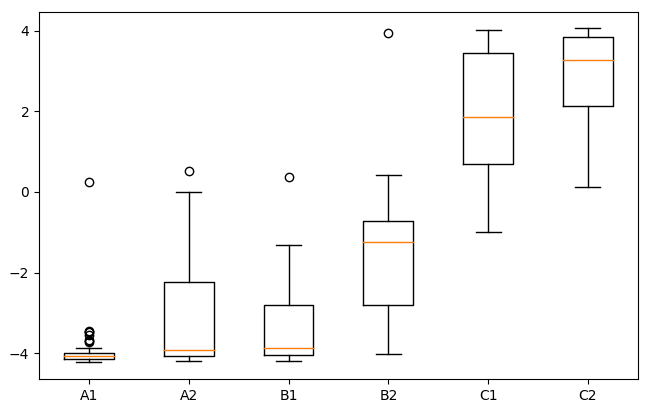}
    \caption{Adaptive simulation on real learner data, using manually judged CEFR level of
      each question as its difficulty estimate ($b_{i}$).  X-axis---6 CEFR levels; Y-axis---ability estimate.}
    \label{fig:simu_level}
  \end{minipage}\hfill
  \begin{minipage}{0.49\linewidth}
    \center
    \includegraphics[trim=4mm 0mm 5mm 0mm, clip, width=\columnwidth]{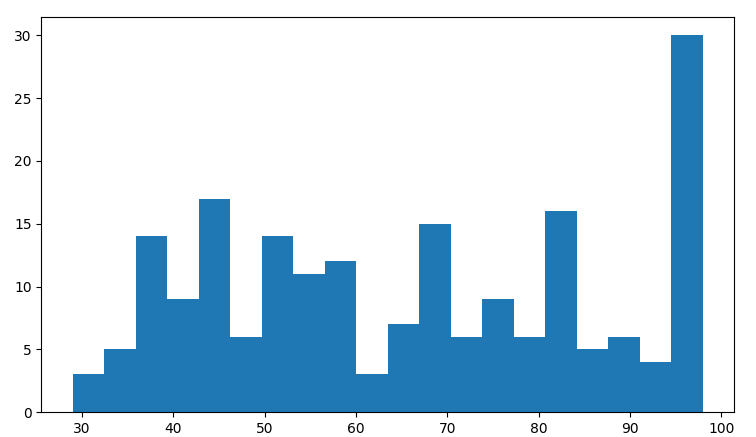}
    \caption{Distribution of length of adaptive test in real learner simulations in
      Figure~\ref{fig:simu_level}.  X-axis---length of test; Y-axis---number of tests in each bin.}
    \label{fig:simu_level_distrib}
  \end{minipage}
\end{figure*}

\subsubsection{Estimation of question difficulty}

The next simulation compares the quality of A.~the parameters of question items learned from
learner data---specifically the difficulty level $b_{i}$ of each item---against
B.~the experts' judgements of the difficulty of the items.
Each question was judged individually on the CEFR scale by expert teachers, as mentioned in
Section~\ref{section:data}.

In this simulation, rather than using the parameters learned by IRT from data---discrimination $a$ and difficulty $b$---we apply $a_{i}=1$ uniformly, and replace $b_{i}$ with the experts' judgement of the difficulty of question $i$.

The expert judgements of difficulty are mapped from the CEFR scores to the difficulty scale as follows.  We take the range of the difficulty values $b$ learned from data for all items---from the minimum to the maximum---and split this range of difficulties evenly into 6 bins of equal width.  The CEFR level of the question (assigned manually by experts) is mapped to the center of each of the 6 bins on the difficulty scale.

Figures~\ref{fig:simu_level} and~\ref{fig:simu_level_distrib} show the results of this simulation.
The examinees' true CEFR levels correspond much less to ability predicted by using ``manual'' estimates of difficulty, compared to Figure~\ref{fig:simu_real},
%
where the means of the boxes increase linearly with ability.
%
%
%
Also, Figure~\ref{fig:simu_level_distrib} shows that the mean length of the test is much higher, uniformly distributed, and many tests fall into the rightmost bin, which is the {\em maximum} number of iterations allowed in this simulation---these tests were forced to terminate at 100 iterations, and may not have converged.

\begin{figure*}
  \begin{minipage}{0.49\linewidth}
    \centering
    \center
    \includegraphics[trim=4mm 0mm 2mm 0mm, clip, width=\columnwidth]{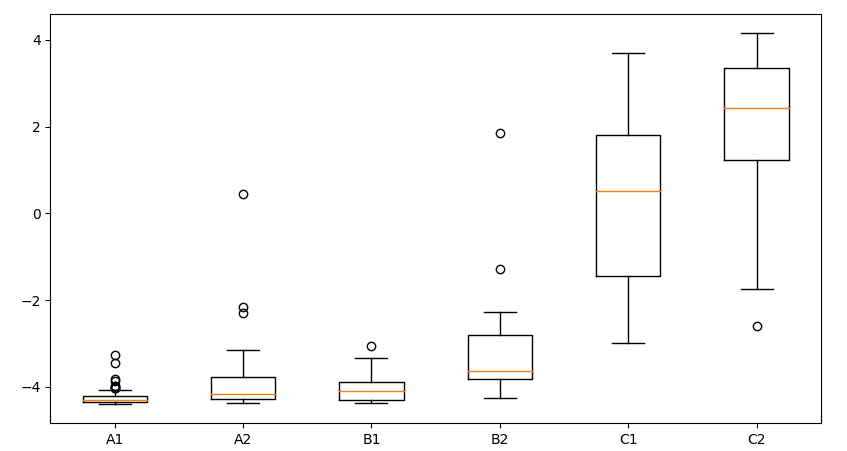}
    \caption{Exhaustive test on real learners with manually labelled CEFR levels. X-axis---the 6 CEFR levels; Y-axis---ability estimate.}
    \label{fig:profile_test_all}
  \end{minipage}\hfill
  \begin{minipage}{0.49\linewidth}
    \center
    \includegraphics[trim=4mm 0mm 2mm 0mm, clip, width=\columnwidth]{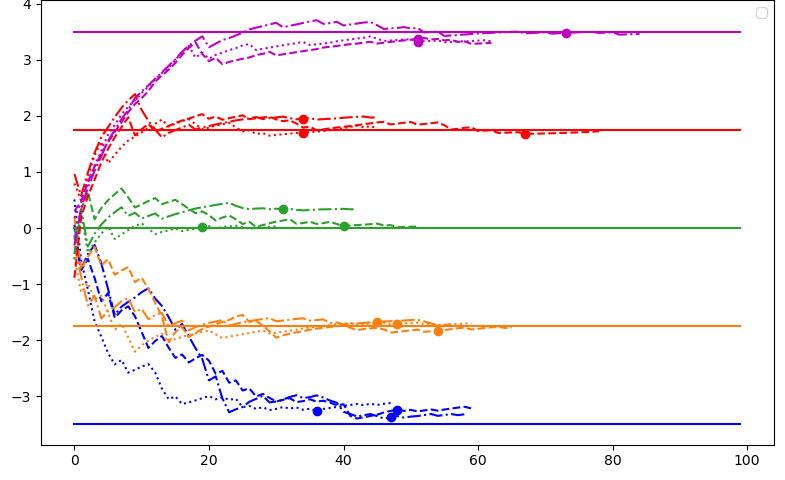}
    \caption{Simulation with ``artificial'' learners: 5 ``true'' levels of proficiency.  3
      learners sampled at each level.  X-axis---number of questions; Y-axis---estimate of ability.}
    \label{fig:simu_artificial}
  \end{minipage}
\end{figure*}

Lastly, to give the ``full benefit of the doubt'' to the exhaustive test, we apply the IRT model to the entire session (containing answers to 300 questions), rather than using item selection according to Equation~\ref{eq:info}, and visualize the result in Figure~\ref{fig:profile_test_all}.
The result is similar to Figure~\ref{fig:simu_level}.  
The correspondence is still far worse than in Figure~\ref{fig:simu_real}.
This indicates that the question levels labeled by experts are not nearly as accurate as the item parameters learned from data ({\bf RQ3}).

Further, as mentioned, the size of the exhaustive test was 300---five times longer than most of the adaptive test sessions need to converge.
Figures~\ref{fig:profile_test_all}, \ref{fig:simu_level} and \ref{fig:simu_level_distrib} show that although the exhaustive test process was imperfect, IRT was still able to use learner data collected from that process, and the model trained on this data is more accurate and more efficient than the exhaustive test ({\bf RQ2}). 

\subsubsection{Slip and exploration in adaptive test}
\label{section:slip_exp}

As mentioned in Section~\ref{section:slip_exp-introduction}, the IRT-based adaptive test is affected by random slips and exploration.  We investigate the exploration parameter settings in Appendix~\ref{sec:appendixI-slip}.

Figure~\ref{fig:simu_artificial} shows simulations conducted with ``artificial'' learners, free of slip and exploration.  At each iteration, we sample an item from the top 5
most informative questions according to Equation~\ref{eq:info}.  Each artificial learner is placed at one of 5 ``true'' proficiency levels, which are represented by the solid horizontal lines; each true level is represented by its own color.  On each level, we simulate 3 artificial learners---represented by a dotted line, dashed line, and dash-dot line.  In these simulations, we model convergence by the EarlyStop termination criterion with hyper-parameters $N = 10$ steps and $\delta = 0.05$ (see Section~\ref{section:termination-criteria}).  To observe the progress of the simulation with this criterion, we continue each simulation for additional 10 iterations  {\em beyond convergence}, with the point of convergence marked by a dot on each simulated line.

As seen from Figure~\ref{fig:simu_artificial}, all 15 simulations converge quite near to their respective targets---to the solid lines.  Most simulations converge well before 60 steps; a few simulations converged between 60 and 80 steps---a much more favorable length, compared to the original 300.

In practice, convergence of the adaptive test is strongly affected by aberrant responses {\em early on} in the test. Therefore, we explore specifically simulations with early aberrant responses.  We increase the slip rate to $\epsilon_{slip}=0.6$ at the beginning, to simulate frequent early aberrant responses, and restore to normal $\epsilon_{slip}=0.1$ after the first 10 questions.

\begin{figure*}
  \begin{minipage}{0.49\linewidth}
    \center 
    \includegraphics[trim=4mm 0mm 3mm 2mm, clip, width=\columnwidth, height=0.63\columnwidth]{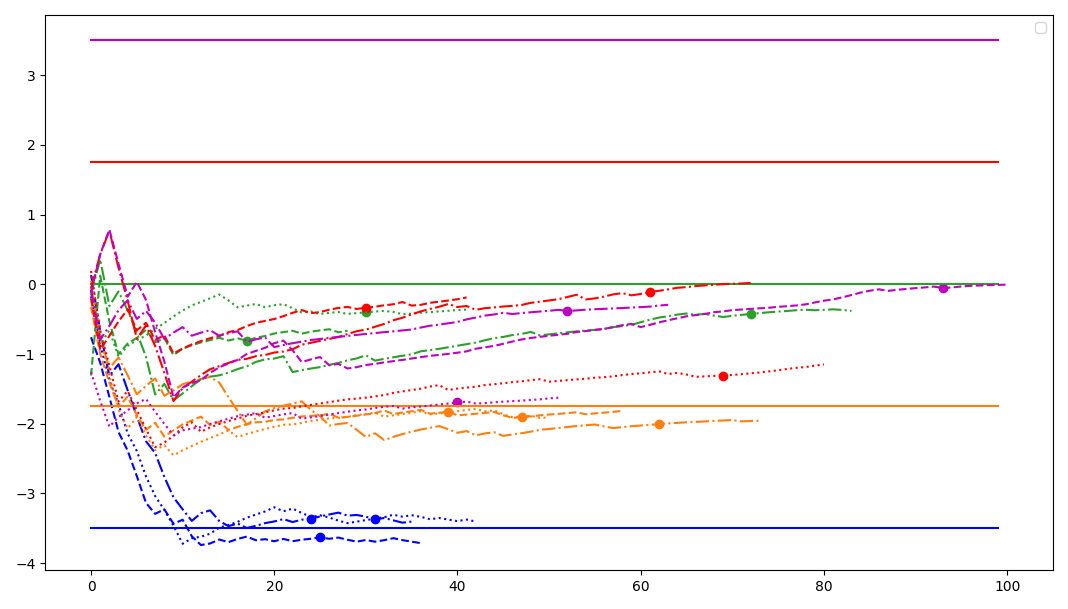}
    \caption{Early slip, no warm-up. Simulation with ``artificial'' learners: 5 ``true'' levels of proficiency, 3 learners sampled at each level.  X-axis---number of questions; Y-axis---estimate of ability.}
    \label{fig:simu_artificial_early_aberrant}
  \end{minipage}
  \hfill
  \begin{minipage}{0.49\linewidth}
    \centering
    \center 
    \includegraphics[trim=4mm 0mm 3mm 2mm, clip, width=\columnwidth, height=0.63\columnwidth]{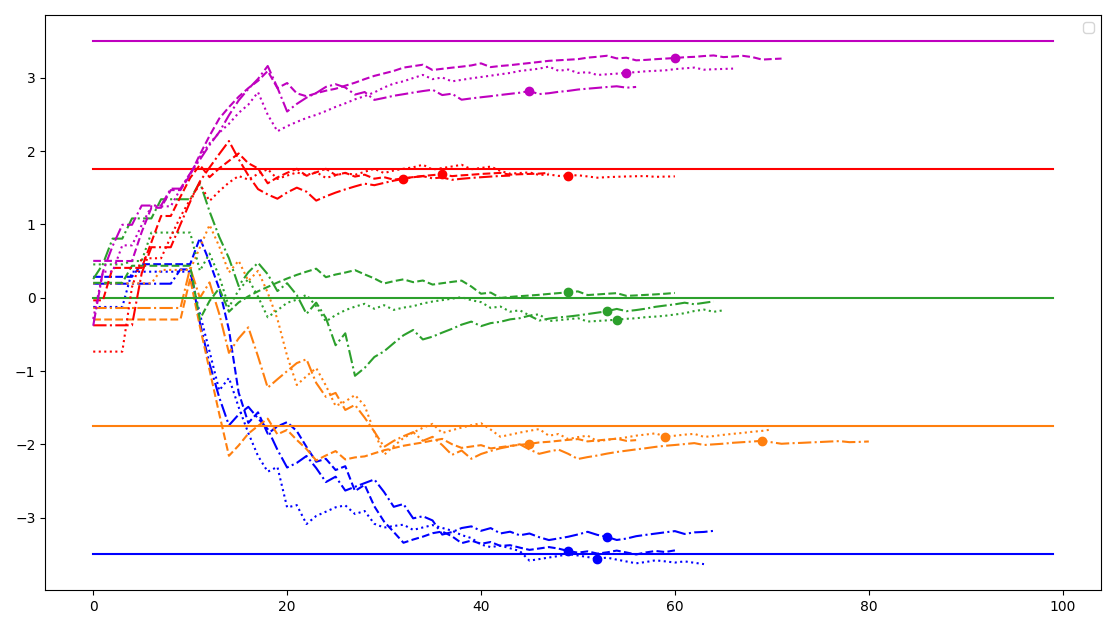}
    \caption{Early slip, with warm-up. Simulation with ``artificial'' learners: 5 ``true'' levels of proficiency, 3 learners sampled at each level.  X-axis---number of questions; Y-axis---estimate of ability.}
    \label{fig:simu_artificial_warm_up}
  \end{minipage}
\end{figure*}

The result can be seen in Figure~\ref{fig:simu_artificial_early_aberrant}.
Early aberrant responses have an asymmetric effect on the final ability estimate. For high-ability students, simulations converge to significantly underestimated scores.  For low-ability students, simulations produce a relatively accurate estimate.

To compensate for this effect, we introduce a ``warm-up'' phase---initial wrong responses are discarded and are not fed into the model.  A simulation with a warm-up phase of 10 steps, with the same setup as Figure~\ref{fig:simu_artificial_early_aberrant}, is shown in Figure~\ref{fig:simu_artificial_warm_up}.
Although the warm-up phase leads to early overestimation for low-ability students, it significantly reduces the effect of early aberrant responses overall.  All simulations eventually converge to an accurate estimate. Meanwhile, the length of the test is 60 on average, which is still about the same size as simulations in Figure~\ref{fig:simu_artificial}.
  
A discussion of termination criteria for the adaptive test can be found in Appendix~\ref{sec:appendixII-termination}.

\subsection{Experiments with exercise data}


\begin{figure*}[t]
  \begin{subfigure}{.5\textwidth}
    \centering
    \includegraphics[trim=5mm 12mm 5mm 0mm, clip, width=.8\linewidth]{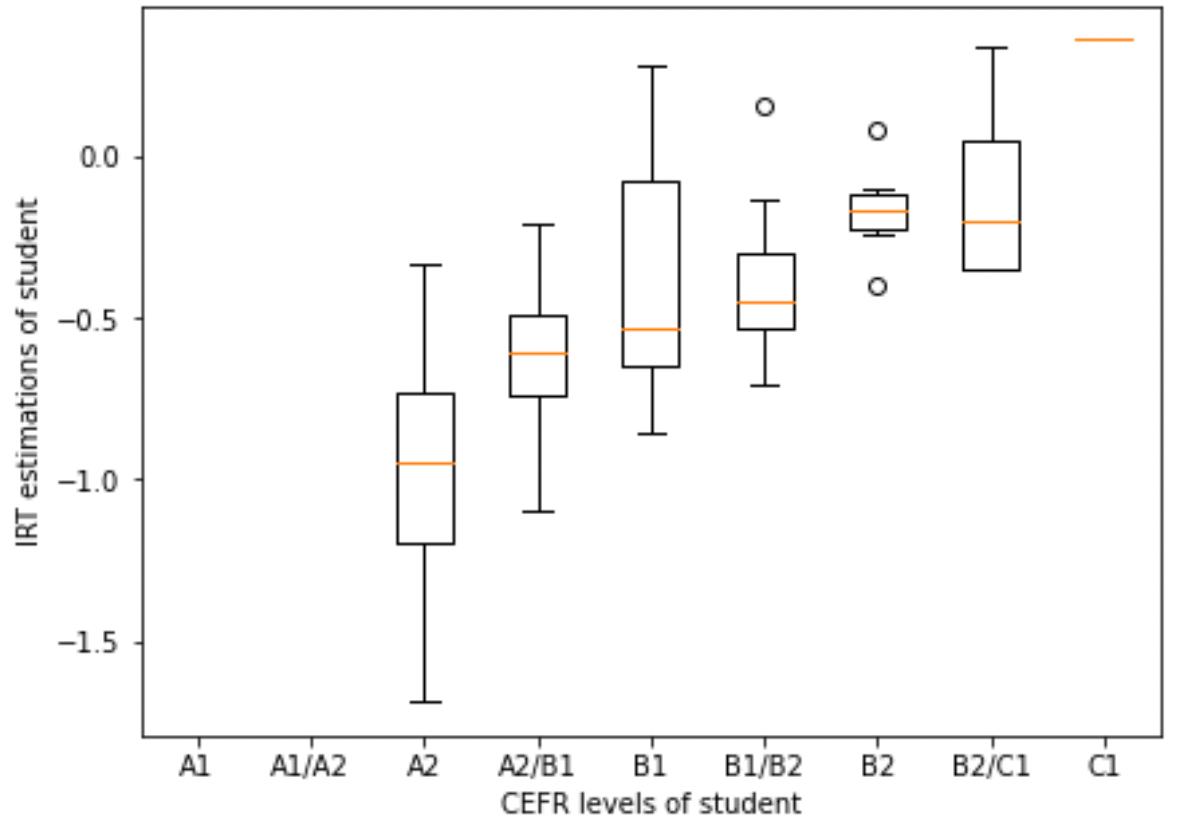}
    \caption{$min_{exer} = 50$, $min_{constr} = 1$, $\rho = .62$}
    \label{fig:50-1}
  \end{subfigure}%
  \begin{subfigure}{.5\textwidth}
    \centering
    \includegraphics[trim=2mm 12mm 5mm 0mm, clip, width=.8\linewidth]{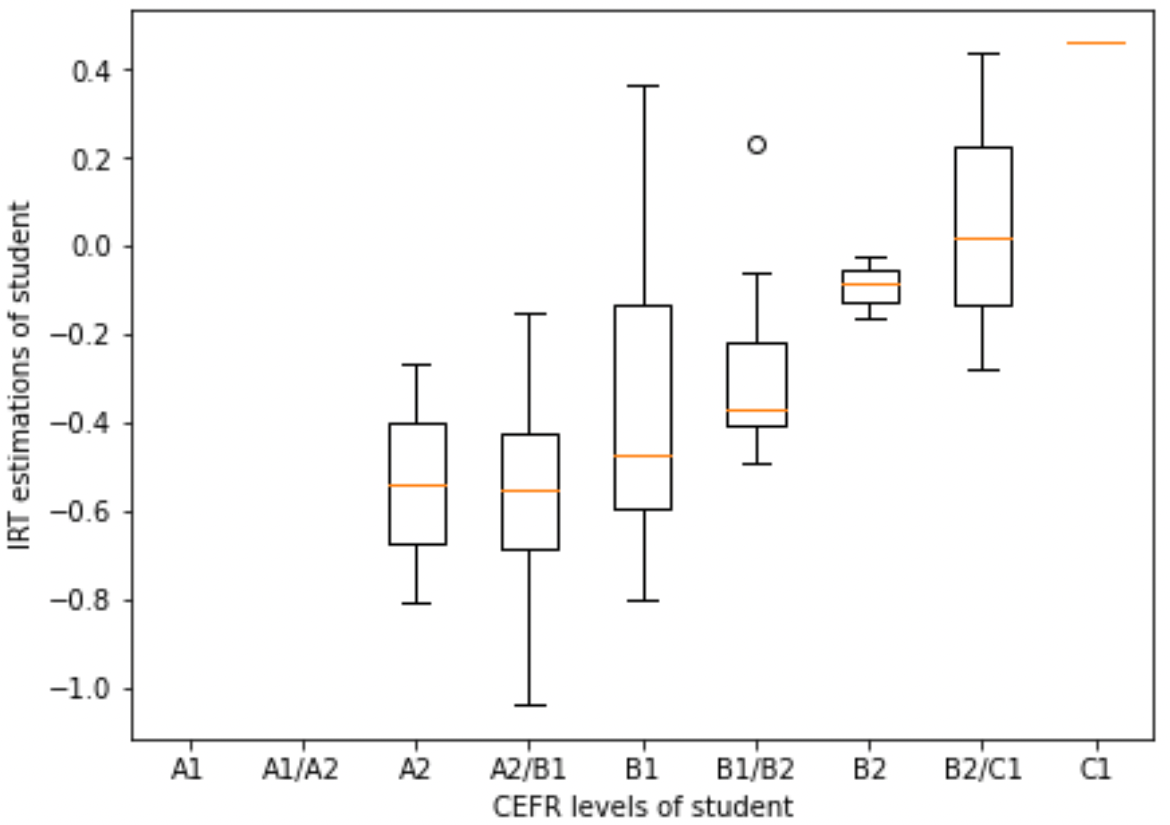}
    \caption{$min_{exer} = 100$, $min_{constr} = 1$, $\rho = .608$}
    \label{fig:100-1}
  \end{subfigure}
  \begin{subfigure}{.5\textwidth}
    \centering
    \includegraphics[trim=5mm 12mm 5mm 0mm, clip, width=.8\linewidth]{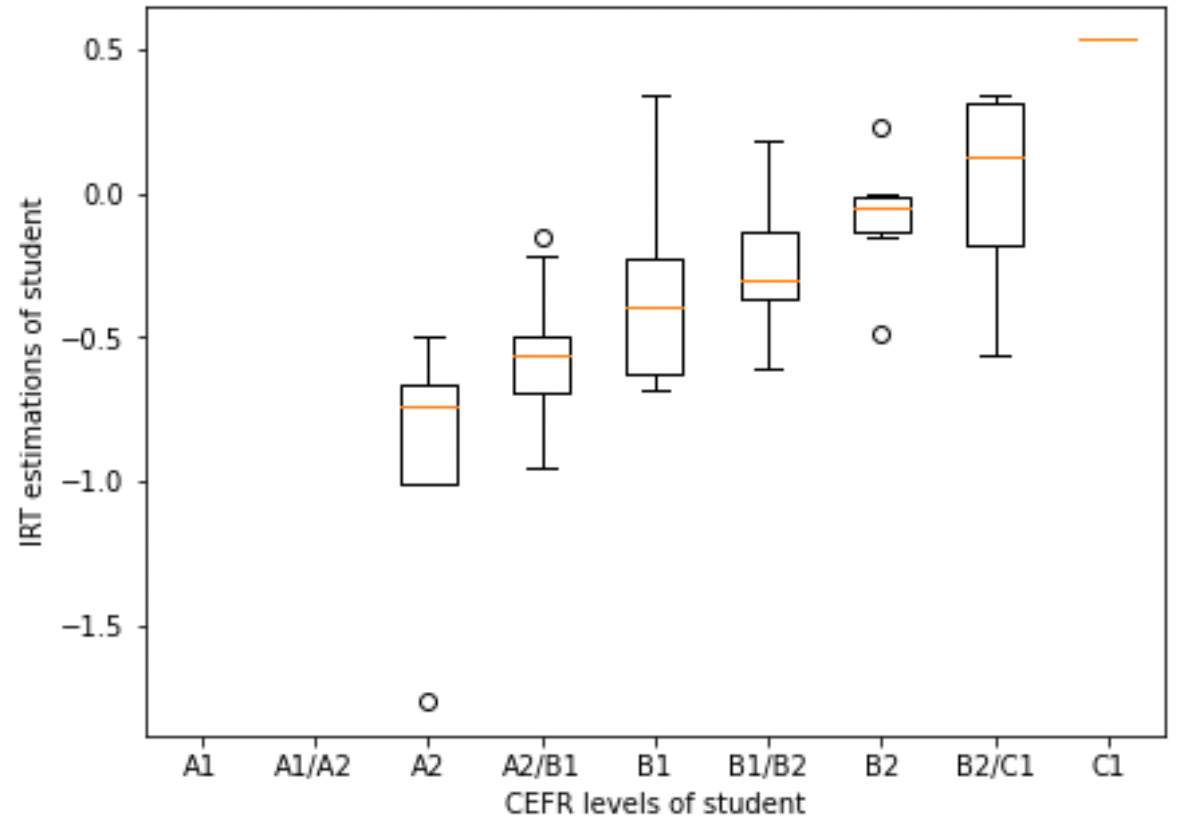}
    \caption{$min_{exer} = 50$, $min_{constr} = 4$, $\rho = .663$}
    \label{fig:50-4}
  \end{subfigure}%
  \begin{subfigure}{.5\textwidth}
    \centering
    \includegraphics[trim=5mm 12mm 4mm 0mm, clip, width=.8\linewidth]{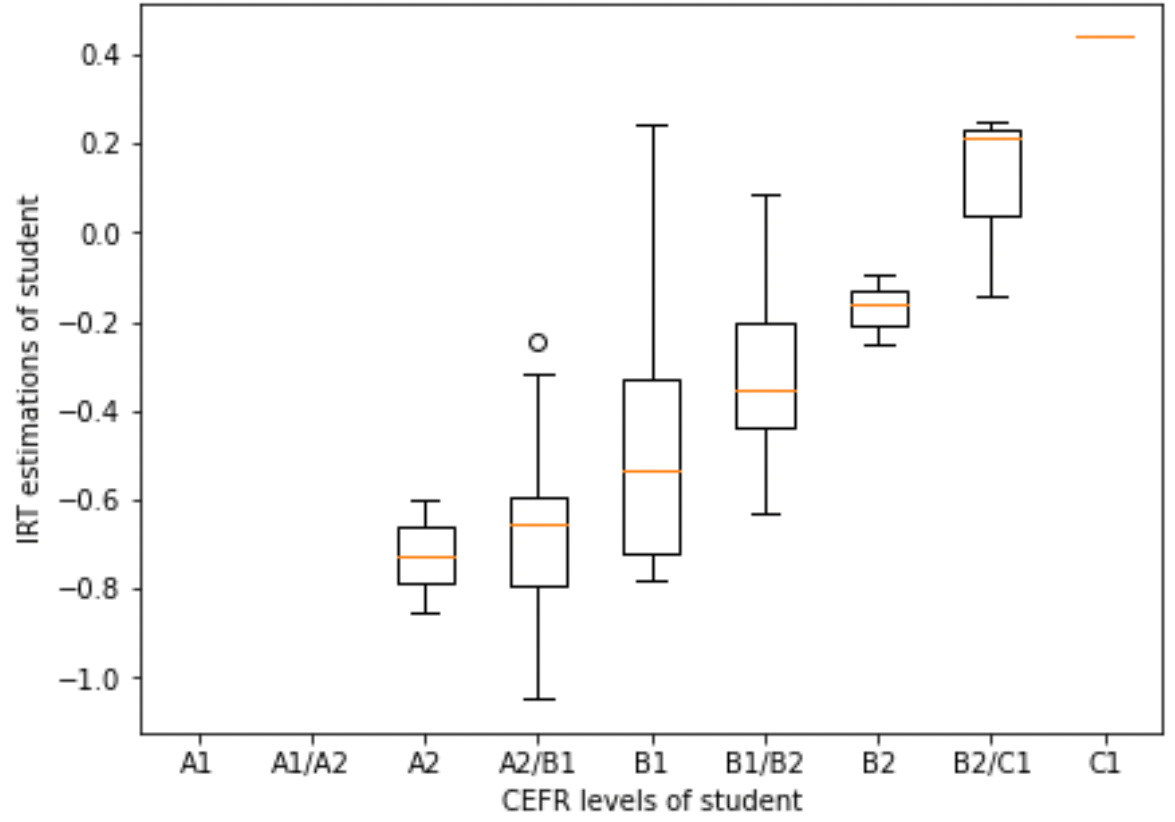}
    \caption{$min_{exer} = 100$, $min_{constr} = 4$, $\rho = .724$}
    \label{fig:100-4}
  \end{subfigure}
  \begin{subfigure}{.5\textwidth}
    \centering
    \includegraphics[trim=2mm 15mm 5mm 0mm, clip, width=.8\linewidth]{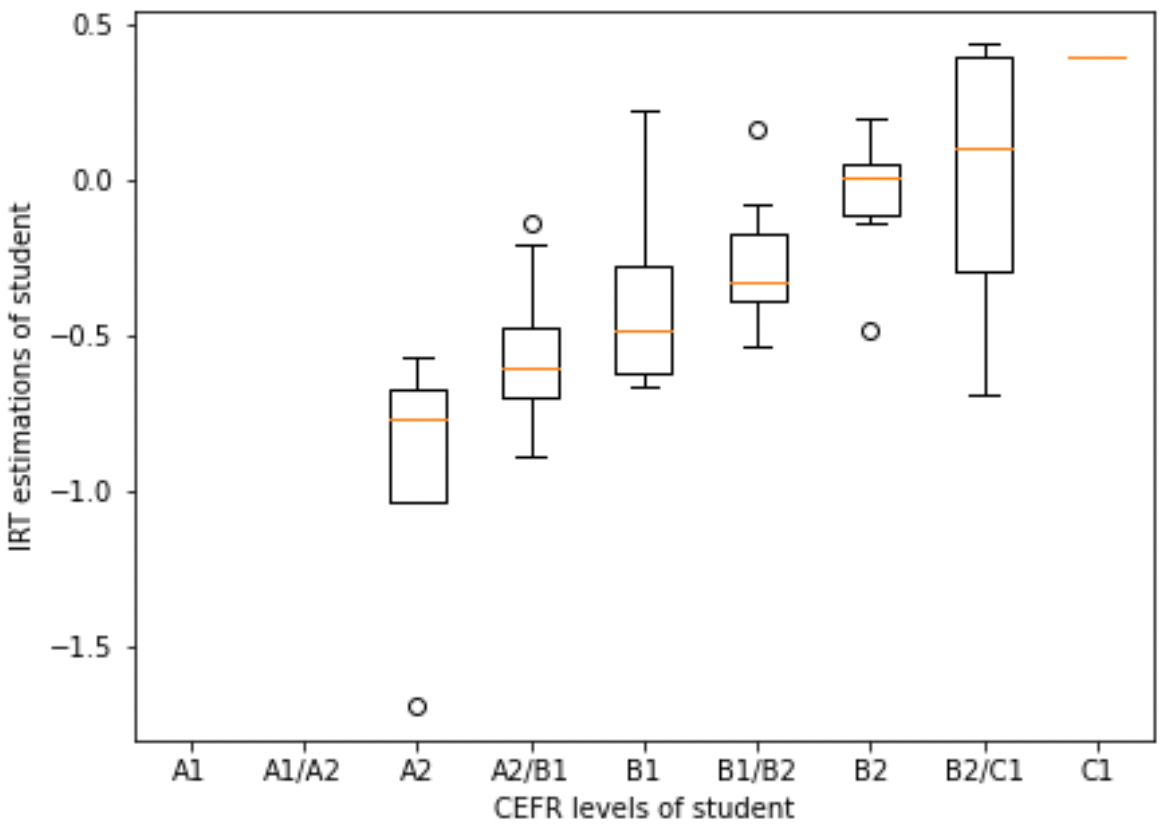}
    \caption{$min_{exer} = 50$, $min_{constr} = 7$, $\rho = .68$}
    \label{fig:50-7}
  \end{subfigure}%
  \begin{subfigure}{.5\textwidth}
    \centering
    \includegraphics[trim=2mm 15mm 5mm 0mm, clip, width=.8\linewidth]{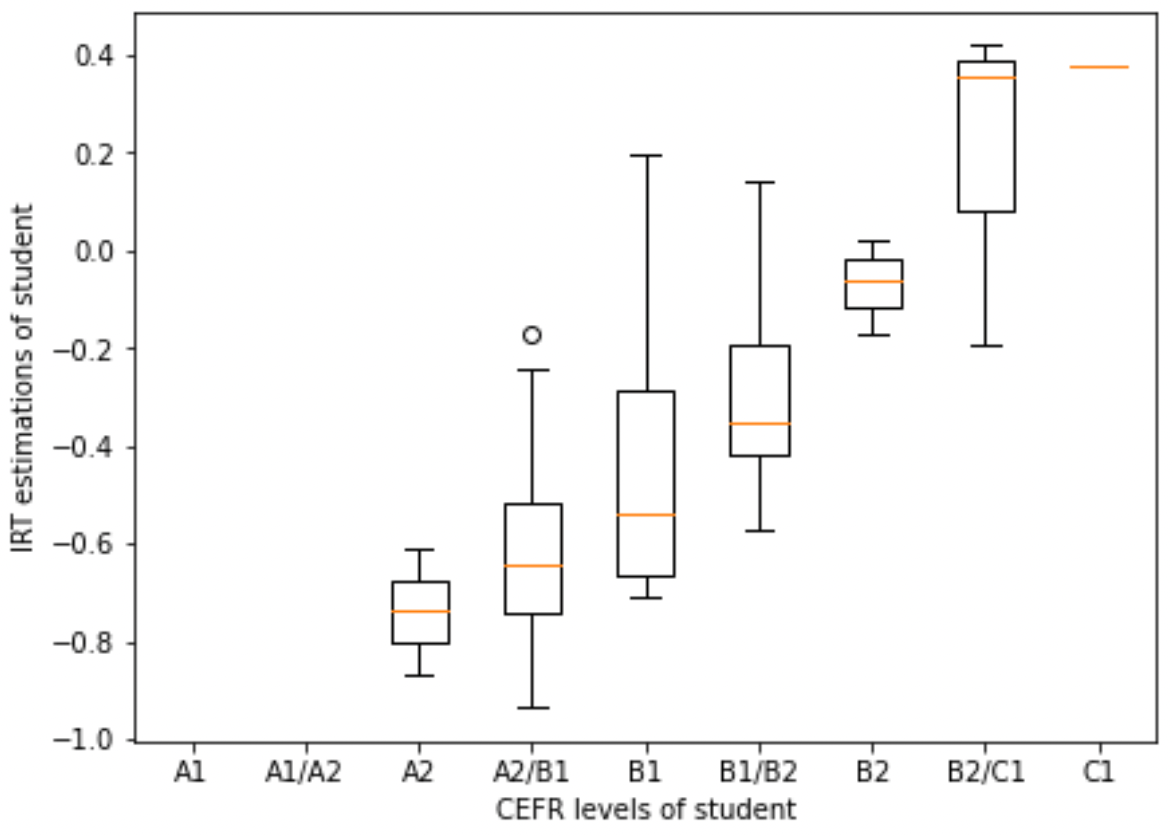}
    \caption{$min_{exer} = 100$, $min_{constr} = 7$, $\rho = .75$}
    \label{fig:100-7}
  \end{subfigure}
  \caption{Estimations of student ability: X-axis---teacher estimates; Y-axis---IRT estimates.}
  \label{fig:irt_exe_corr_plot}
\end{figure*}

We turn to the final research question: Can we model learner ability reliably based on the learners' responses to {\em exercises}, without explicit testing?

The setup of experiments with exercises is described in Section~\ref{sec:exercise-analytics}.
We study how the quality of assessment is affected by two factors: A.~the amount of exercises done by the student, and B.~the amount of answers collected per construct. 
For A: we train the IRT model only on learner data from students who have done a substantial number of exercises; we consider students who have done only a few exercises unreliable for collecting statistics. The parameter $min_{exer}$ is the minimum number exercises that must be done by student $S$ in order to use data from $S$.
For B: once the IRT model has been trained, we estimate the ability of each student $S$ using her data only for those constructs for which $S$ has answered at least $min_{constr}$  exercises. (We assume that if $S$ has done too few exercises with a construct $C$, her statistics for that constructs are less reliable.)

\COMMENT{In Figures over 50 or over 100 exercises, which is approximately 1000 students. 
To evaluate the IRT estimation, we compare the estimated IRT scores with the CEFR levels assessed by teachers of 50 students.
}

The plots in Figure~\ref{fig:irt_exe_corr_plot} show how ``real'' CEFR estimates (from teachers) compare to estimated ability scores from IRT based on their exercise performance. The x-axes represent the ``real'' CEFR and the y-axes represent the estimated ability scores. For each CEFR score, the figures show the distribution of IRT estimated scores as box plots for students corresponding to that CEFR score. 
The figures on the left side use  $min_{exer} = 50$, whereas those on the right side use  $min_{exer} = 100$.

For each box plot, the orange line indicates the median and the box indicates the interquartile range which includes $50\%$ ($25\%$ fall between the first quartile and median, $25\%$ fall between the third quartile and median) of the students of that level. Outside the interquartile are the whiskers which include approximately $50\%$ of the students. Outside the whiskers are the outliers.

With $min_{exer} = 50$, we can use data from 1316 students; the evaluation uses 52 students for whom the teachers have given CEFR scores.
Raising $min_{exer}$ to 100, we can use data only from 982 students---but these students have done more work---and the evaluation uses 43 of these students (with CEFR scores).

As expected, training IRT on data from students who have done more exercises shows stronger correlation with the teachers' grades.
Also, as expected, as we estimate a student's ability based on constructs for which she has seen more exercises, we also obtain better correlation: from $min_{constr} = 1$ in the top row to $min_{constr} = 7$ in the bottom row.

The best performance, in Figure~\ref{fig:100-7} ($\rho = 0.75$), is based on a model trained on data from students who have done the most exercises, and using data from constructs seen more frequently.

The key overall observations drawn from the experiments with exercise data:
\begin{itemizerCompact}

\item The ability of the tutor to discriminate between students by CEFR level based on their exercise data in Figure~\ref{fig:100-7} is at least as good as the ability to discriminate based on adaptive test scores, in Figure~\ref{fig:simu_real}, ({\bf RQ4}).

\item The more the students practice during their coursework, the higher the values of $min_{exer}$ and $min_{constr}$ we can fix, leading to better quality of ability prediction.

\item Overall, we can expect to be able to collect much more learner data from exercises than from tests, since the tests are not a learning activity {\em per se}, whereas the students will be motivated to provide more practice data over time---provided that the tutor offers interesting content and a good learning environment.

\end{itemizerCompact}


\section{Conclusions and Future Work}
\label{section:conclusion}

In order to support {\em personalized} tutoring, one of the most essential tasks of ITS is to access the learner's proficiency.
We present the application of Item Response Theory, to assess students in two contexts: test and exercise.

Our aim is for a more efficient and accurate assessment on student's proficiency, so the tutor can offer better items during tests, and better exercises during practice, based on the assessment.
\paragraph{Assessment of ability through testing:} Our goal is to build an adaptive test as a component of the ITS (to replace the exhaustive testing used previously), that not only gives accurate estimates of the examinee's ability, but is also quicker. 

We have introduced our data sources and the bank of test items.
Then, we introduce an approach to fitting an IRT model, and simulation procedures.  The IRT model has randomness, including random initialization, random item selection, slip and exploration; we discussed the motivation for randomness in the testing process.

The experiments in the context of testing show:
\begin{enumeratorCompact}
  \item In spite of an imperfect initial data collection process, the parameters learned from the learner data, are usable to train an IRT model that results in a viable adaptive test.
  \item Models that use difficulty estimates learned from data far outperform the models that use difficulty estimates assigned to items by pedagogy experts. This agrees with the findings of other researchers, e.g.,~\cite{lebedeva-2016-placement-test}.

  \item Slips generally increase the length of a test.  However, introducing an Exploration step is able to slightly compensate for the delay in convergence caused by slips.
    
  \item The IRT model can give an accurate prediction at a very early stage of a test. Additionally, more questions will generally allow the IRT model to achieve a better prediction.
\end{enumeratorCompact}
We expect that over time the adaptive tests will yield more learner data: since the test does not stress the learners, they will be more willing to use it periodically to check their ability levels, voluntarily.  This will enable us to collect better learner data, and we expect to build more accurate and efficient IRT models on that data.

\paragraph{Assessment of ability through exercise practice:}
To our knowledge, this is the first application of IRT to learner data collected from {\em exercise practice} in language learning---where we have no explicit fixed item bank, and we must estimate the difficulty of {\em latent} variables (not directly observable), which capture the performance of the students on particular linguistic constructs---linked to the exercises in a one-to-many fashion.

Note, that we do not propose to {\em replace} explicit testing altogether with implicit assessment during practice.
Explicit testing is a valuable instrument in language learning, which provides essential information about the learner's state.
Rather, this work suggests that:
(A) Testing can be reserved for use at special junctures in the learning process, for example, for summative assessment at the end of a course; ``placement'' testing, initially; etc.
For continual assessment over time, as the learner does many exercises---they provide adequate information to form an accurate, detailed assessment.

(B) When explicit testing {\em is} used, we can resort to adaptive rather than exhaustive testing, which provides an equally accurate assessment.

\bibliography{custom, ../BIB/thebib-long, ../BIB/revita}
\section*{Statements and Declarations}
\subsection*{Funding}
This work was supported in part by the Academy of Finland,
Helsinki Institute for Information Technology (HIIT),
BusinessFinland (Grant 42560/31/2020), 
and 
Future Development Fund, Faculty of Arts, University of Helsinki.
\subsection*{Competing Interests}
The authors have no relevant financial or non-financial interests to disclose.

\subsection*{Authors' contributions}
All authors contributed to the system implementation and experiments. Data collection and analysis were performed by [Jue Hou], [Anh-Duc Vu] and [Anisia Katinskaia]. The first draft of the manuscript was written by [Jue Hou] and [Roman Yangarber].  All authors commented on previous versions of the manuscript. All authors contributed to the final manuscript.

\begin{appendices}

\section{Experiments with slip and exploration}
\label{sec:appendixI-slip}

In this section, we investigate how the test is affected by slip, and whether exploration can compensate for the effects caused by slip.  We apply the Metrics introduced in Section~\ref{section:method}, and conduct 500 simulations. Simulations are conducted at random ability levels $\theta_{true}$ uniformly sampled from $[-3.5, 3.5]$.

\begin{figure*}
\begin{minipage}{0.49\linewidth}
\centering
    \center 
    \includegraphics[trim=1mm 1mm 0mm 1mm, clip, width=\columnwidth]{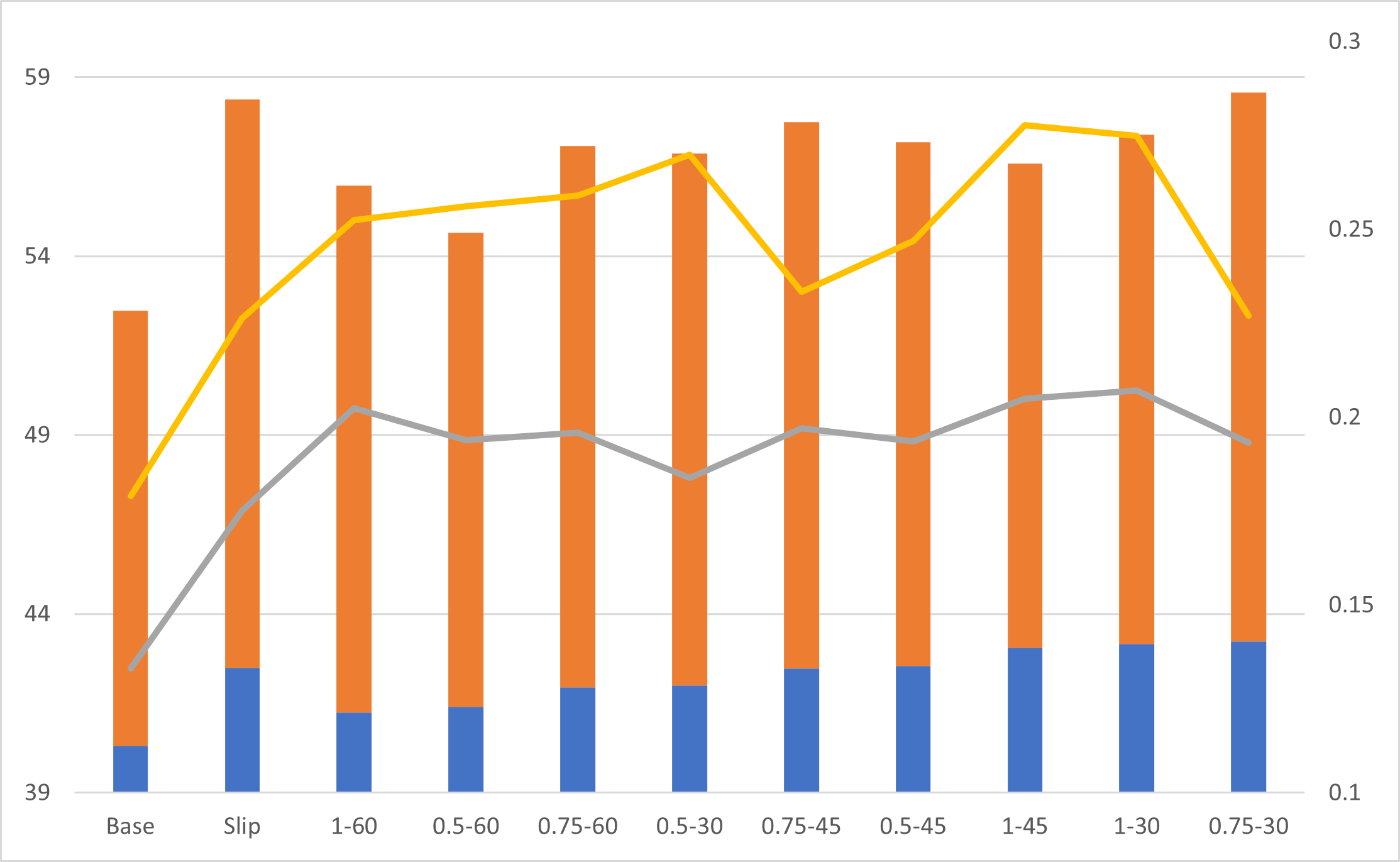}
    \caption{Simulations of slip and exploration. }
    \label{fig:exp_slip}
\end{minipage}\hfill
\begin{minipage}{0.49\linewidth}
    \center 
    \includegraphics[trim=1mm 1mm 0mm 1mm, clip, width=\columnwidth]{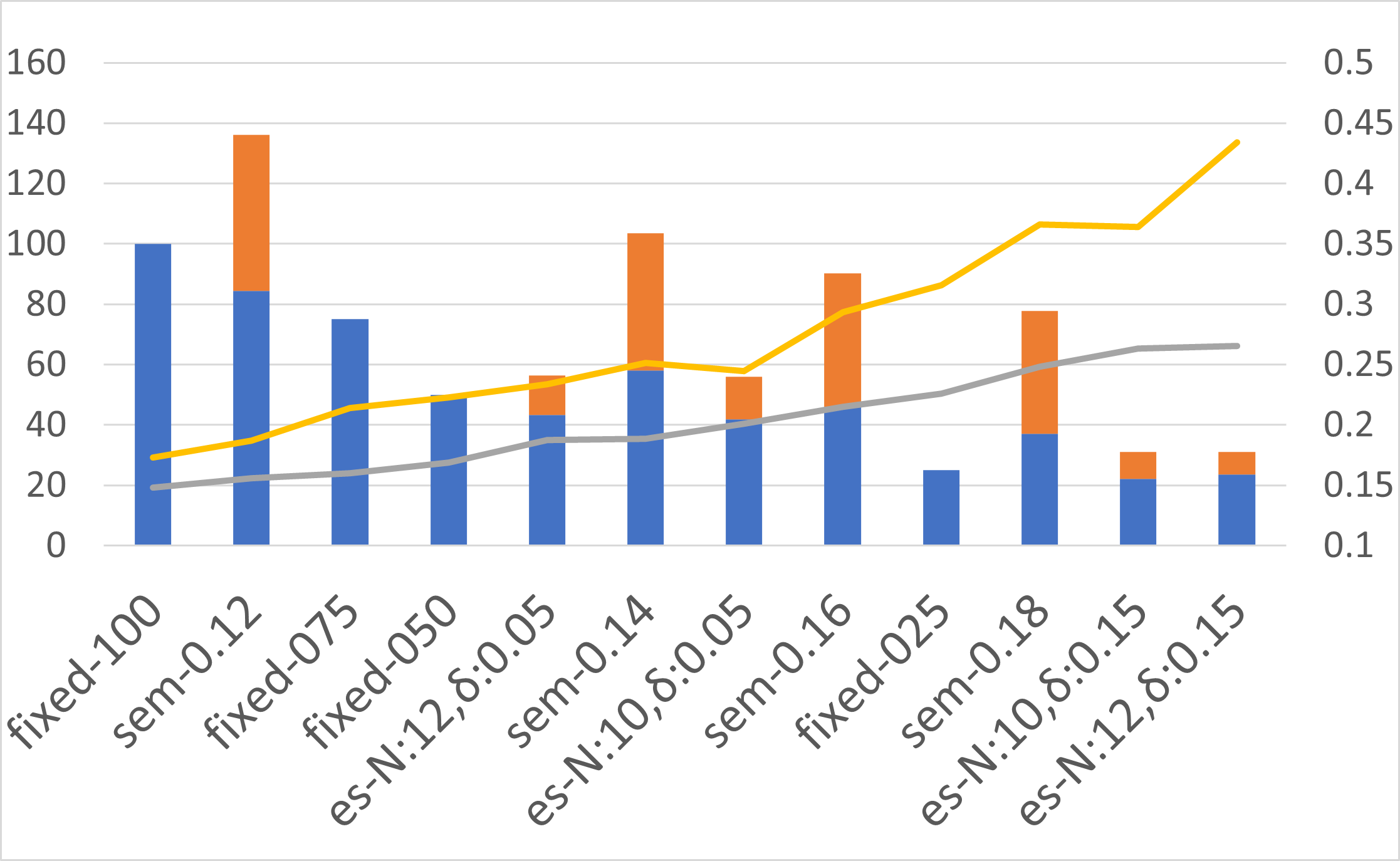}
    \caption{Overall comparison among all criteria.}
    \label{fig:overall_term}
\end{minipage}
\end{figure*}

Figure~\ref{fig:exp_slip} shows the performance of various settings for exploration parameters.  The first two labels on the X-axis indicate the baseline performance and performance with slip but no exploration.
The rest of each label consists of two settings: the first number indicates the exploration factor $\alpha$,
while the second one indicates the exploration range $N_{exp}$.  The bar chart corresponds to the Y-axis on the left, which shows the number of iterations.
Each bar describes the distribution of the length of the simulations.  The blue part on the bottom of each bar indicates the mean length of the simulations, while the orange bar at the top indicates the SD of the length of the simulations.

The lines correspond the Y-axis on the right using the same scale as $\theta$---predicted ability.  The gray line shows the MAE of the prediction of ability, and the yellow line shows the SD of the error.

We can make several observations from Figure~\ref{fig:exp_slip}:
\begin{itemizerCompact}

\item Slips generally increase the number of iterations needed to converge, as well as the error between $\theta_{true}$ and the predicted $\theta$.  This is expected because the IRT model will increase the number of questions to compensate for the ``instability'' of learner responses.

\item Although exploration is not able to compensate for the error caused by slipping, it helps to converge faster with slightly fewer iterations, and much fewer iterations in the worst case.

\end{itemizerCompact}

\section{Experiments with Termination Criteria}
\label{sec:appendixII-termination}

In this section, we explore the efficiency and effectiveness of the adaptive test procedures by conducting simulations with various termination criteria.  The simulations are conducted in a similar way as in Section~\ref{section:slip_exp}.  Each simulation in this section is conducted with only one single randomly chosen $\theta_{true}$, rather than at several different ability levels, as done in the preceding sections.

We conducted three simulations for the three types of termination criteria, and one overall simulation.  For each evaluation, we run each simulation 500 times and uniformly sample $\theta_{true}$ from $[-3.5, 3.5]$,
which is the same as in Section~\ref{section:slip_exp}. We use the $\alpha=0.5$ and the exploration parameter $N_{exp}=60$.

Figures~\ref{fig:sem_term}, \ref{fig:earlystop_term} and \ref{fig:overall_term} in this section follow the same visualization scheme.  The X-axis represents the type of termination criteria with their corresponding parameters.  The bar charts follow the Y-axis on the left side of the figure, and show the number of iterations until convergence.  The blue part at the bottom of each bar represents the mean number of iterations, while the orange part at the top represents the SD of the number of iterations.

The line graphs follow the Y-axis on the right side of the plot, on the same scale as the ability scores $\theta$.
The gray and yellow lines represent MAE and SD of error, respectively.


More detailed experiments with termination criteria can be found in Appendix~\ref{sec:appendixIII-stopping}.
Here, we discuss the overall comparison.  We select the best parameter setting from each type of stopping criterion: 
\begin{itemizerCompact}
\item Fixed-length: 25, 50, 75, 100
\item SEM: 0.12, 0.14, 0.16, 0.18
\item EarlyStop: ($N$:10, $\delta$:0.05), ($N$:10, $\delta$:0.15), ($N$:12, $\delta$:0.05), ($N$:12, $\delta$:0.15)
\end{itemizerCompact}
Figure~\ref{fig:overall_term} shows the performance of these stopping criteria.  The simulations are sorted according to MAE (grey line).  As we can see, the MAE of these criteria are relatively close, between 0.14 and 0.27, which suggests an error rate below 3.9\% in terms of the observed range of ability $\theta$ ($-3.5 \leqslant \theta \leqslant +3.5$). Comparing the test size and SD of error, we observe that more iterations generally allow the IRT model to achieve a better prediction.


\section{Stopping criteria in detail}
\label{sec:appendixIII-stopping}

\paragraph{Fixed-length:} As the baseline for all termination criteria, we first explore the simulation with a fixed-length test.  We check lengths from 25 to 150 questions, with an increment of 25.

\begin{figure}[t]
\center
\includegraphics[trim=1mm 1mm 0mm 1mm, clip, width=0.5\columnwidth]{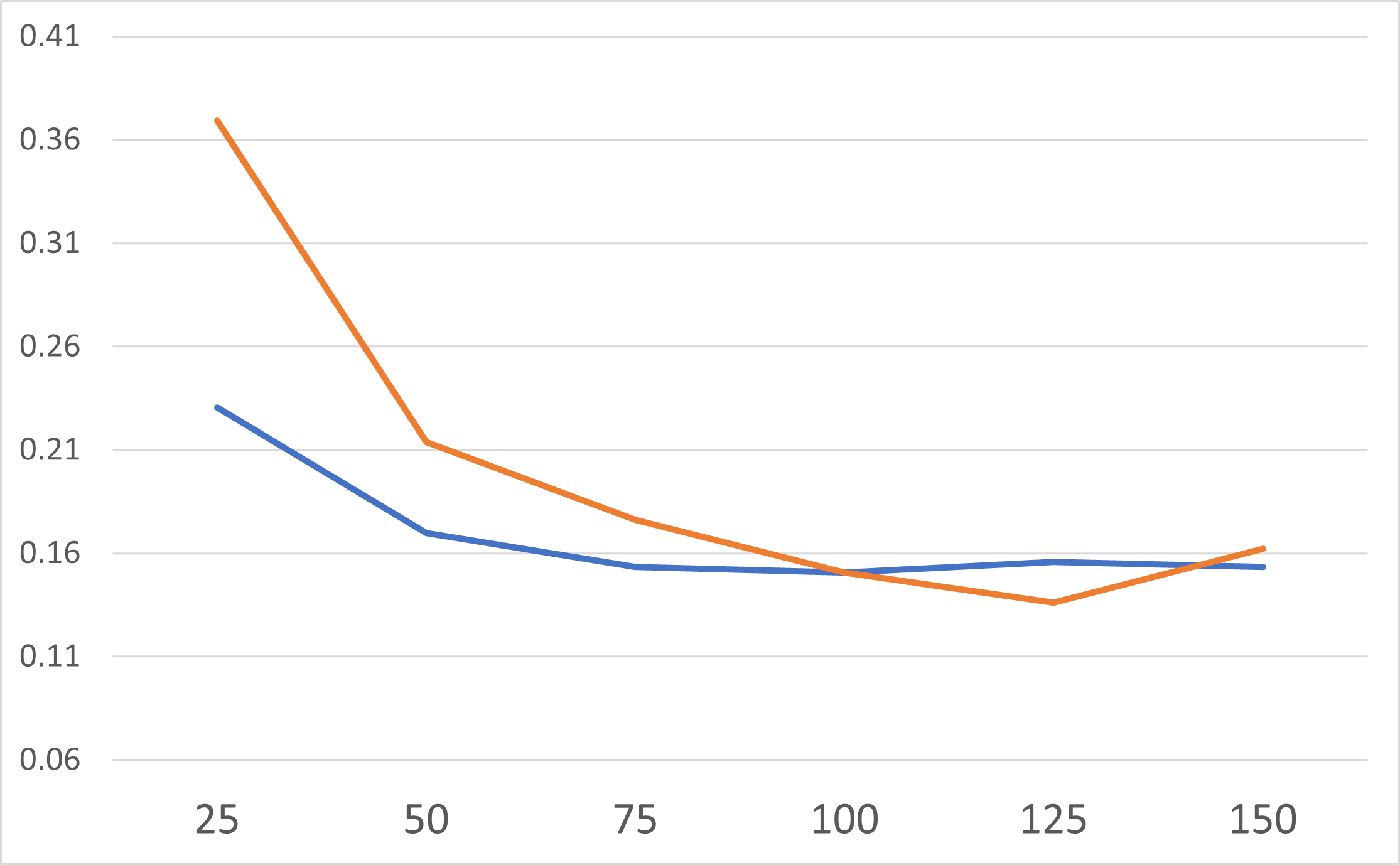}
\caption{Performance of fixed-length termination criterion.}
\label{fig:fixed_term}
\end{figure}

Figure~\ref{fig:fixed_term} shows the accuracy for these test lengths.  The blue line indicates MAE, while the orange line indicates the SD of error.


The Figure shows that a fixed-length test with $50$ or more iterations generally has MAE below $0.17$ on the ability scale.
This is quite adequate, considering that the range of observed ability is 6 units, from $-3.5$ to $+3.5$ (in Figure~\ref{fig:simu_real}); MAE value of $0.17$ constitutes $2.4\%$ of this range.
\COMMENT{This is quite adequate, if one CEFR level occupies about $.66$ ability units.}
The SD of error decreases as the number of iterations increases.  This is as expected, because more iterations leads to more observations, which will allow the IRT model to give a more reliable estimate.

The worst MAE is observed when running with only 25 iterations, which results in an error rate of $0.23$ on the ability scale.
This suggests that the IRT model may start to converge and give estimate that are close to the true level as early as 25 iterations.
Figure~\ref{fig:fixed_term} also shows that performance remains about the same beyond 100 iterations.  These observations, together with observation from Figure~\ref{fig:simu_artificial}, help us set the lower and upper boundaries for our adaptive tests.


\paragraph{SEM:} We next explore the SEM criterion.  We check it from 0.1 to 0.3 with an increment of 0.02, i.e., from higher to lower confidence.


\begin{figure*}
\begin{minipage}{0.49\linewidth}
\centering
    \center 
    \includegraphics[trim=1mm 1mm 0mm 1mm, clip, width=\columnwidth]{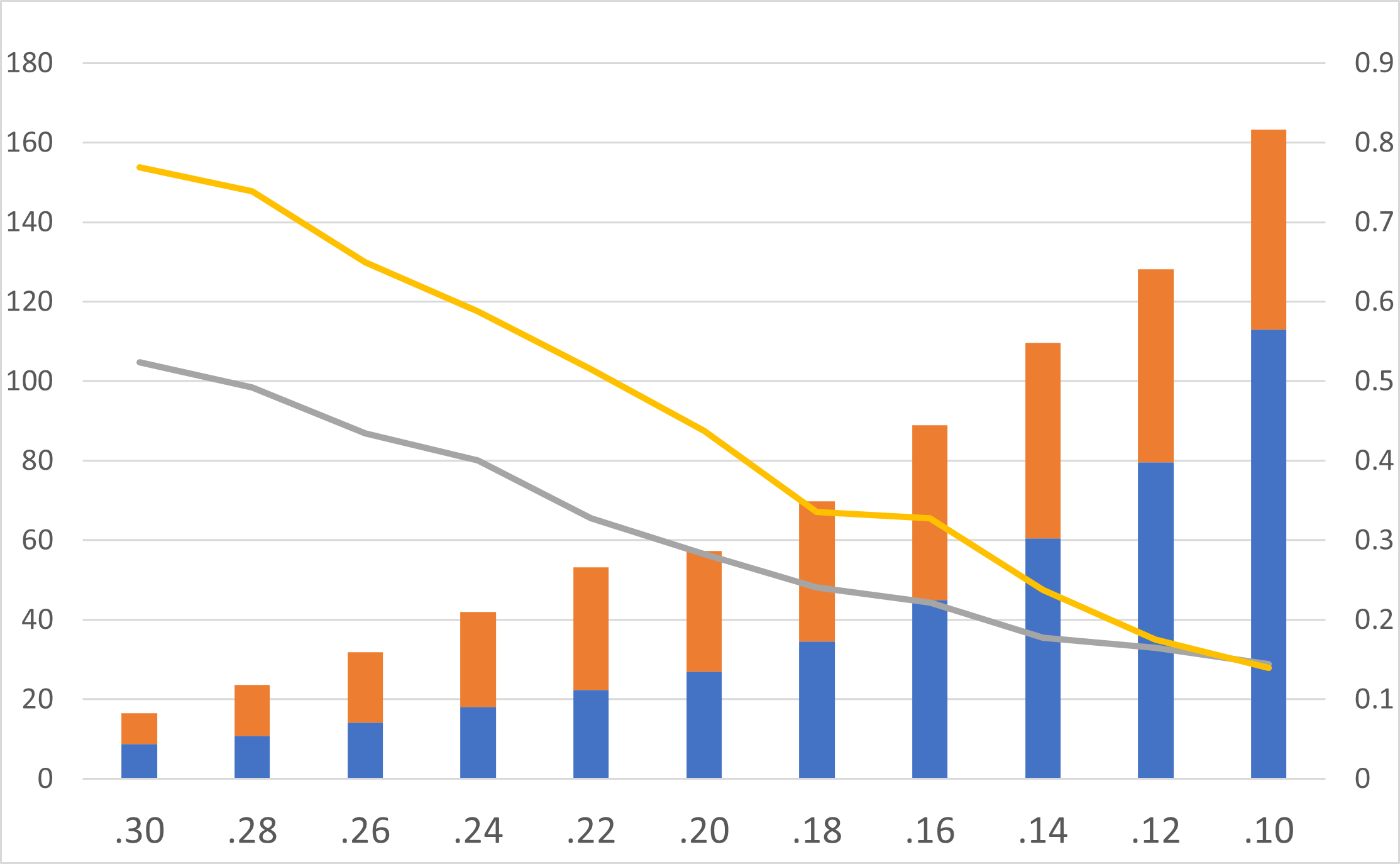}
    \caption{Performance of SEM termination criterion. }
    \label{fig:sem_term}
\end{minipage}\hfill
\begin{minipage}{0.49\linewidth}
    \center 
    \includegraphics[trim=1mm 5mm 1mm 1mm, clip, width=\columnwidth]{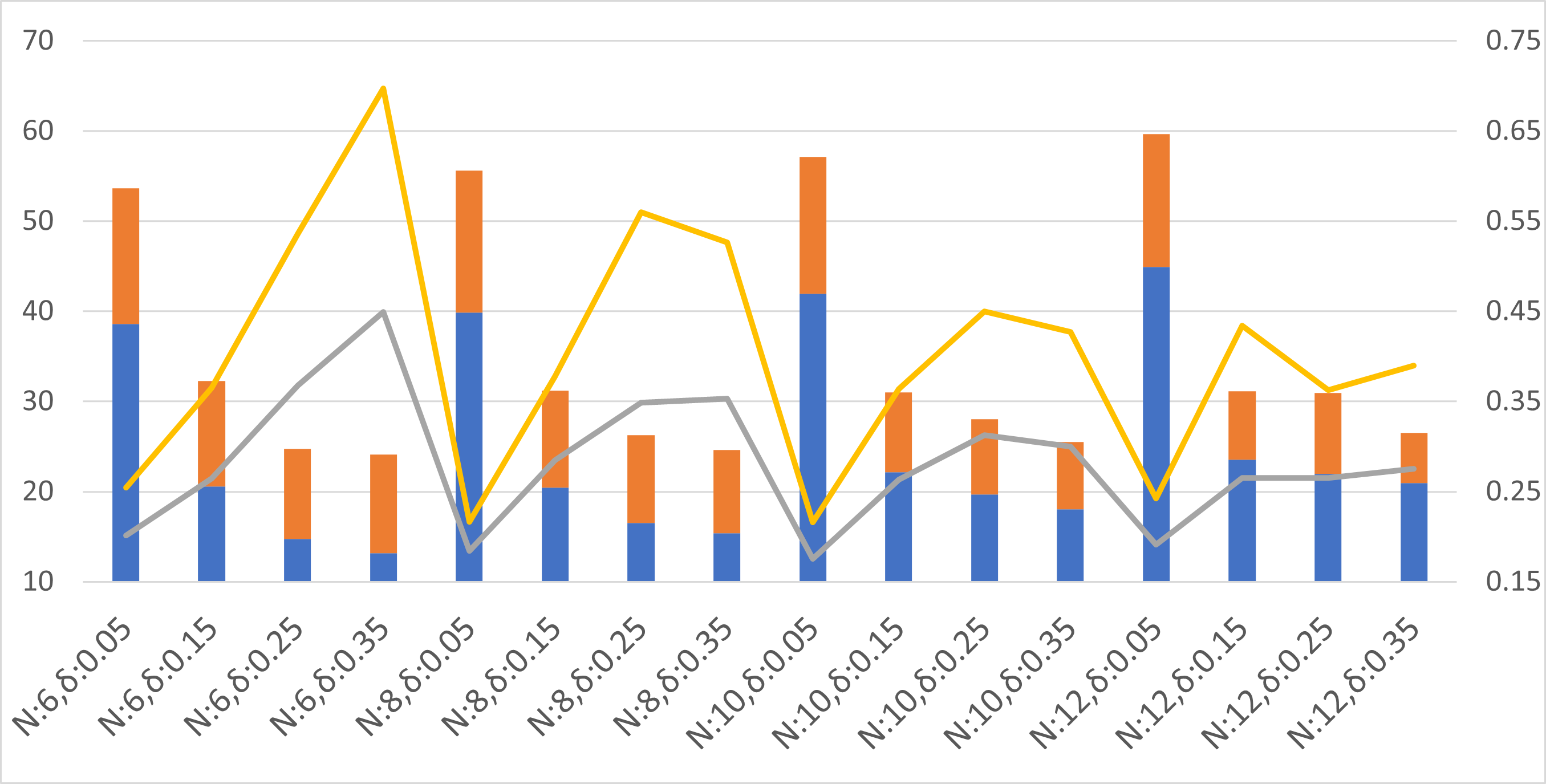}
    \caption{Performance of EarlyStop termination criterion.}
    \label{fig:earlystop_term}
\end{minipage}
\end{figure*}

Figure~\ref{fig:sem_term} shows the different SEM conditions.  The Figure shows that the more constrained the confidence, the lower the MAE and the lower the SD of error.
This is achieved at the expense of the test length required for convergence, which increases with increased confidence.  On the other hand, we can observe the test length is more stable with looser constraint on the left side of Figure~\ref{fig:sem_term}, as the orange bar is small.
This is expected, since with less restriction, the IRT model will converge more easily.
If the constraint is more restricted, it will be more difficult to converge.  Due to the randomness in our simulations, this will lead to more variance in the actual length of the test.  This may explain the higher SD value of the number of iterations when there is more constraint.

\COMMENT{At the same time, the MAE on the left side of the figure shows a greater fluctuation if
compared to its right side.  Together with the high SD of error, this suggests with less
restricted constraint, the accuracy of the ability estimates will vary more strongly.
Taking the test length into consideration, this suggests an insufficient number of
iterations.}



\paragraph{EarlyStop:} We next explore the third stopping criterion.  The idea behind this criterion is to stop the test once the ability value is fluctuating within a small range $\delta$ over the last $N$ iterations.
We vary $\delta$ from 0.05 to 0.35 with an increment of 0.1, while $N$ varies from 6 to 12 with an increment of 2.  Higher values of $N$ indicates a stricter criterion, while lower values of $\delta$ indicates a stricter criterion.  The combination of these two parameters gives 16 simulations, which is visualized in Figure~\ref{fig:earlystop_term}.  The simulations appear sorted by $N$ and $\delta$.

As seen from Figure~\ref{fig:earlystop_term}, except for the worst two, the MAE is mostly below 0.35, which suggests a low error rate of $<5\%$ over the observed range of $\theta$ ($-3.5 \leqslant \theta \leqslant +3.5$, in Figure~\ref{fig:simu_real}).
For a given $N$, we see that the higher $\delta$, the higher the MAE and the SD of error. This is expected because the $\delta$ suggests a less restricted constraint.  The test length required for convergence decreases as $\delta$ becomes higher.
On the other hand, for a given $\delta$, we can observe a similar behaviour as $N$.  A greater value of $N$ means the MAE and the SD of error becomes lower.

At the same time, the SD of test length also decreases with a greater value of $\delta$ or smaller value of $N$. This is a similar situation as in Figure~\ref{fig:sem_term}.  It is easy to converge when the constraint is relaxed, but it will be difficult with more uncertainty with a stricter constraint.



\end{appendices}

\end{document}